\pdfoutput=1

\documentclass[11pt]{article}

\usepackage[final]{acl}

\usepackage{times}
\usepackage{latexsym}

\usepackage[T1]{fontenc}

\usepackage[utf8]{inputenc}

\usepackage{microtype}

\usepackage{inconsolata}

\usepackage{graphicx}
\usepackage{graphicx}
\usepackage{graphicx}
\usepackage{amsmath}
\usepackage{amssymb}  
\usepackage{graphicx}
\usepackage{xspace}
\usepackage{makecell}
\usepackage{adjustbox}
\usepackage{comment}
\usepackage{booktabs}
\usepackage{multirow}
\title{From Retrieval to Generation: Comparing Different Approaches}

\author{
    \textbf{Abdelrahman Abdallah, Jamshid Mozafari, Bhawna Piryani,} \\
    \textbf{Mohammed Ali, Adam Jatowt} \\
    University of Innsbruck \\
    \texttt{\{abdelrahman.abdallah, jamshid.mozafari, bhawna.piryani,} \\
    \texttt{mohammed.ali, adam.jatowt\}@uibk.ac.at}
}

\begin{document}
\maketitle

\begin{abstract}

Knowledge-intensive tasks, particularly open-domain question answering (ODQA), document reranking, and retrieval-augmented language modeling, require a balance between retrieval accuracy and generative flexibility. Traditional retrieval models such as BM25 and Dense Passage Retrieval (DPR) efficiently retrieve from large corpora but often lack semantic depth. Generative models like GPT-4-o provide richer contextual understanding but face challenges in maintaining factual consistency.  In this work, we conduct a systematic evaluation of retrieval-based, generation-based, and hybrid models, with a primary focus on their performance in ODQA and related retrieval-augmented tasks. Our results show that dense retrievers, particularly DPR, achieve strong performance in ODQA with a top-1 accuracy of 50.17\% on NQ, while hybrid models improve nDCG@10 scores on BEIR from 43.42 (BM25) to 52.59, demonstrating their strength in document reranking. Additionally, we analyze language modeling tasks using WikiText-103, showing that retrieval-based approaches like BM25 achieve lower perplexity compared to generative and hybrid methods, highlighting their utility in retrieval-augmented generation.  By providing detailed comparisons and practical insights into the conditions where each approach excels, we aim to facilitate future optimizations in retrieval, reranking, and generative models for ODQA and related knowledge-intensive applications\footnote{The code and the dataset will be available after acceptance of the paper.}.


\end{abstract}

\section{Introduction}


The increasing complexity of knowledge-intensive tasks, particularly open-domain question answering (ODQA) and retrieval-augmented applications, necessitates advanced approaches to efficiently retrieve and generate relevant information. Traditionally, retrieval-based methods have played a central role in these tasks, with models like BM25~\cite{robertson2009probabilistic} serving as foundational tools for extracting relevant documents. However, the limitations of keyword-based retrieval prompted the development of dense retrieval models such as Dense Passage Retrieval (DPR)~\cite{karpukhin2020dense} and Contriever~\cite{izacard2021unsupervised}, which leverage transformer-based architectures to encode queries and documents into dense representations. While dense retrieval models improve over sparse methods, they introduce new challenges. First, retrieval corpora are typically divided into fixed chunks~\cite{karpukhin2020dense}, which can lead to retrieving irrelevant content. Second, dual-encoder architectures encode queries and documents separately, limiting direct interaction between them~\cite{khattab2021relevance}. Finally, dense retrieval models require pre-encoding and storing document embeddings, which constrains scalability and hinders their ability to leverage large language models (LLMs)~\cite{levine2022standing}.


To address these limitations, generative models such as GPT-3.5 and InstructGPT~\cite{brown2020language,ouyang2022training} offer an alternative by directly generating contextualized responses instead of retrieving existing documents. Approaches like GenRead~\cite{yu2022generate} first generate relevant text and then use it for answer prediction. However, generative models often struggle with factual consistency and may hallucinate information~\cite{huang2023survey}, making them less reliable for knowledge-intensive tasks.
Given the trade-offs between retrieval and generation, hybrid models have emerged to integrate the strengths of both approaches. Merging Generator and Retriever (MGR)~\cite{abdallah2023generator,zhang2023merging} combines generated and retrieved documents, allowing models to refine answers while maintaining factual accuracy. However, hybrid models introduce challenges in document selection, requiring effective reranking strategies to prioritize relevant information. Recent work in Retrieval-Augmented Generation (RAG) and In-Context Learning (ICL)~\cite{lewis2020retrieval,ram-etal-2023-context} highlights the importance of reranking techniques in improving answer quality.

\begin{figure}[t]

    \begin{minipage}{0.5\textwidth}
        \centering
        \includegraphics[width=\textwidth]{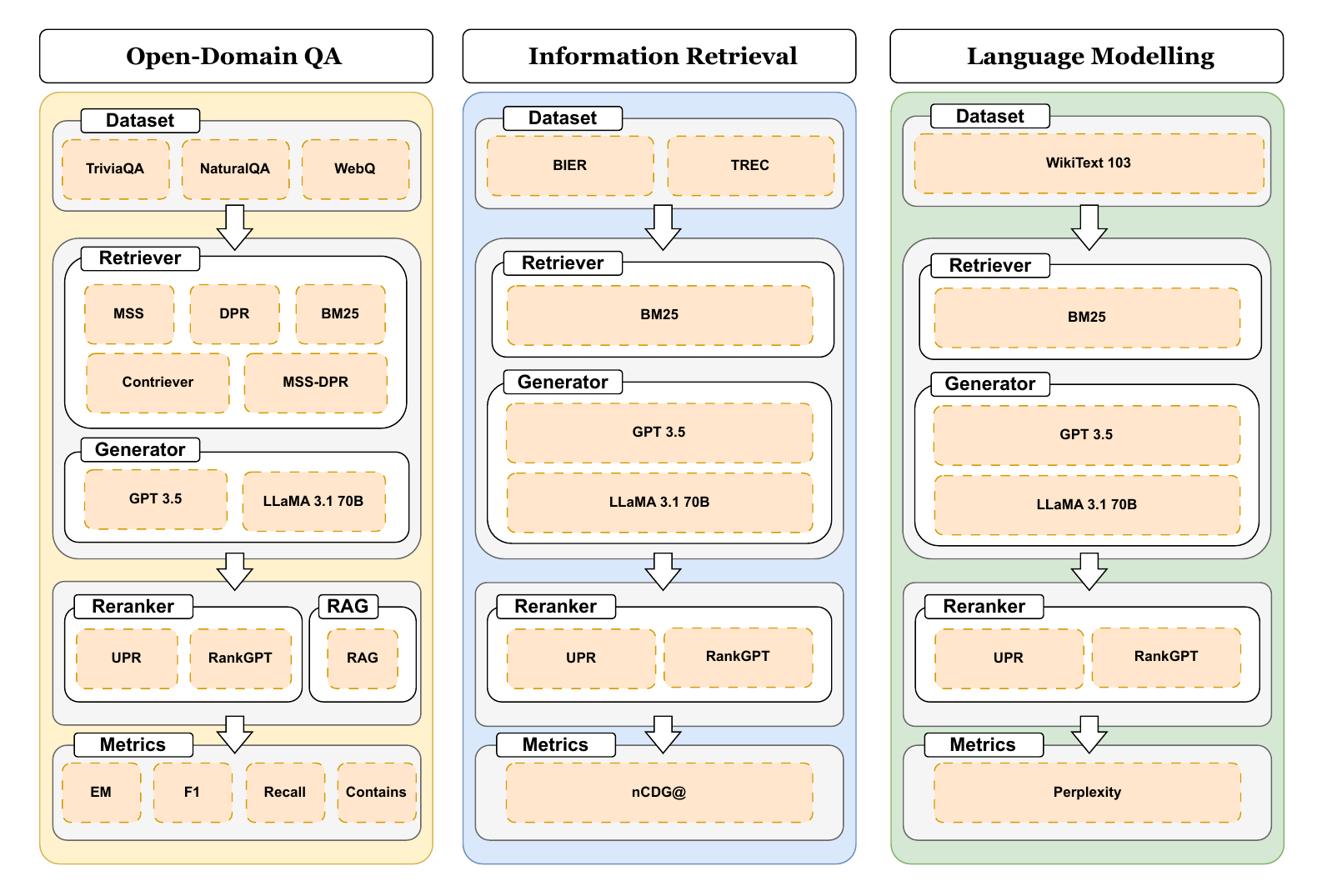}
        \caption{Overview of experimental setup across the three tasks of open-domain question answering (QA), Information Retrieval, and Language Modeling. 
        }
         \label{fig:intro}
    \end{minipage}\hfill
   
\end{figure}

This paper extends prior work on retrieval-augmented language models~\cite{ram-etal-2023-context}, by incorporating generated documents into the retrieval process. We evaluate retrieval, generation, and hybrid models with a primary focus on ODQA and retrieval-augmented tasks, examining their impact on document selection, answer generation, and language modeling. Figure~\ref{fig:intro} provides an overview of these tasks and the experimental setup, illustrating how different approaches are compared in terms of retrieval effectiveness, generative capability, and reranking strategies. First, we compare the accuracy of retrieval models such as BM25~\cite{robertson2009probabilistic}, MSS~\cite{sachan2021end}, MSS-DPR~\cite{sachan2021end}, Contriever~\cite{izacard2021unsupervised}, and DPR~\cite{karpukhin2020dense} with generation-based models. Next, we examine how combining retrieval and generation methods affects performance in hybrid models. We also investigate strategies for reranking documents—an essential step in hybrid models—to determine how best to select the most relevant content for downstream tasks. Additionally, we investigate how hybrid models combining retrieval and generation perform in ODQA and Information Retrieval (IR) tasks, evaluating datasets like BEIR~\cite{thakur2021beir} and TREC~\cite{craswell2020overview}. Finally, we evaluate the impact of these methods on language modelling in reducing perplexity. In general, we try to answer a question: \emph{Which approach—retriever, generator, or hybrid—is best suited for ODQA, language modeling and information retrieval tasks?} Our contributions are as follows:

\begin{enumerate}
    \item We provide a comparison of retrieval and generation-based models, focusing on their effectiveness in ODQA and retrieval-augmented applications by incorporating generated documents into the retrieval process. 
    
    \item  We evaluate the impact of combining retrieval and generation methods in hybrid models, examining how these models perform across different tasks.
    
    \item We explore advanced document reranking methods, demonstrating how reranking enhances Information Retrieval (IR) accuracy and improves the performance of hybrid models in ODQA tasks.

    \item We provide practical insights into the strengths and limitations of retrieval, generation, and hybrid approaches in ODQA and retrieval-augmented applications
\end{enumerate}

\section{Related Work}
Advancements in Open-Domain Question Answering (ODQA), Document Reranking, and Language Modeling (LM) have led to three main approaches: retriever-based, generator-based, and hybrid models, along with retrieval-augmented techniques for improving factual consistency in text generation.

Retriever-based methods focus on identifying relevant documents before processing. Sparse models like BM25~\cite{robertson2009probabilistic} rely on lexical matching, while dense retrieval methods such as DPR~\cite{karpukhin2020dense} and Contriever~\cite{izacard2021unsupervised} encode queries and documents into dense vectors for improved retrieval. Further refinements include ANCE~\cite{xiong2020approximate}, ColBERT~\cite{khattab2021relevance}, and MSS-DPR~\cite{sachan2021end}. Some models bypass full-document retrieval, using dense phrase retrieval~\cite{lee2020learning,seo2019real} to extract answer spans directly.

Generator-based models generate responses rather than retrieving documents. LLMs like GPT-3.5 and InstructGPT~\cite{brown2020language,ouyang2022training} perform well in open-ended QA but struggle with factual accuracy, often hallucinating information~\cite{huang2023survey}. Models like GenRead~\cite{yu2022generate} and DocGen~\cite{askari-etal-2023-expand} generate contextual documents before extracting answers. Pretrained models like T5-RC~\cite{raffel2020exploring} and BART-based retriever-reader architectures~\cite{lewis2020retrieval} attempt to improve reliability, but consistency issues persist.

Hybrid approaches integrate retrieval and generation, balancing factual grounding with contextual flexibility. RAG~\cite{lewis2020retrieval} conditions generation on retrieved documents, while FiD~\cite{izacard2020leveraging} enhances multi-document conditioning. Merging Generator and Retriever (MGR)~\cite{abdallah2023generator,zhang2023merging} dynamically selects between retrieved and generated content. Neural rerankers like MonoBERT~\cite{nogueira2019passage} and DuoBERT~\cite{nogueira2020document} refine retrieval results. Hybrid models have also been effective in In-Context Learning (ICL)~\cite{ram-etal-2023-context}, allowing LLMs to retrieve external knowledge dynamically.

Retrieval-augmented language models improve factual consistency by conditioning text generation on retrieved knowledge. $k$NN-LM~\cite{knn-lm} enhances predictions using nearest-neighbor retrieval but faces scalability challenges~\cite{he-etal-2021-efficient,retomaton}. Retrieve-and-Read models like REALM~\cite{realm} and RETRO~\cite{retro} integrate retrieval into masked modeling or cross-attention to improve factual reliability. These approaches bridge the gap between traditional LMs and knowledge-grounded systems, enhancing generated response accuracy.
\section{Approaches}

\subsection{Retriever Models}
Let $\mathcal{D} = {\mathbf{d}_1, \mathbf{d}_2, \dots, \mathbf{d}_M}$ be a collection of evidence documents representing a retrieval corpus. Given a query $\mathbf{q}$, an Information Retrieval (IR) model selects a subset of relevant passages $\mathcal{Z} \subset \mathcal{D}$, one or more of which will ideally contain the correct answer to $\mathbf{q}$. Our setup supports passages obtained from any retriever, whether based on sparse representations like BM25 ~\cite{robertson2009probabilistic} or dense representations such as DPR~\cite{karpukhin2020dense}, MSS~\cite{sachan2021end}, MSS-DPR~\cite{sachan2021end}, and Contriever~\cite{izacard2021unsupervised}.  BM25 is a traditional sparse retriever that ranks documents based on term frequency-inverse document frequency (TF-IDF).  DPR encodes queries and documents into dense vector representations using pre-trained transformers. The similarity between a query $\mathbf{q}$ and a document $\mathbf{d}$ is calculated as the dot product of their dense embeddings, i.e., $\text{sim}(\mathbf{q}, \mathbf{d}) = E_Q(\mathbf{q})^\top E_P(\mathbf{d})$, where $E_Q$ and $E_P$ are the encoders for the query and document, respectively. In addition to DPR, we test models such as MSS, which focuses on masked salient span prediction, and MSS-DPR, which extends DPR with additional pre-training using MSS. Another dense retriever, Contriever, is trained in an unsupervised manner using contrastive learning on text paragraphs. We assume that each retriever provides the top-K most relevant passages, denoted as $\mathcal{Z} = \{\mathbf{z}_1, \mathbf{z}_2, \dots, \mathbf{z}_K\}$.

\subsection{Generator Model}
The generator approach is an alternative to the traditional retriever. Rather than relying on retrieving documents from an external corpus, it prompts an LLM to generate contextual documents based on a query.  Formally, given a question $\mathbf{q}$, the model generates a set of contextual documents $\mathcal{G} = \{\mathbf{g}_1, \mathbf{g}_2, \dots, \mathbf{g}_n\}$, where each document $\mathbf{g}_i$ is generated by a large language model, such as InstructGPT~\cite{ouyang2022training}, conditioned on the query $\mathbf{q}$. 

\subsection{Zero-Shot LLM Reranking}
\label{sec:method-zero-shot-reranking}

Zero-shot retrieval reranking~\cite{abdallah2025asrank,abdallah2025dynrank} aims to reorder retrieved or generated documents within an Information Retrieval (IR) pipeline, without relying on training data specific to the task or dataset. We explore two methods for reranking: using LLMs to generate a question from a document and a zero-shot ranking approach based on semantic matching.

In the first approach, inspired by the Unsupervised Passage Re-ranking (UPR)~\cite{sachan2022improving}, a pre-trained LLM estimates the likelihood of generating a question given the passage. Formally, for each passage $\boldsymbol{z}_i$ from the set of top-K retrieved passages $\mathcal{Z}$, we compute the relevance score $p(\boldsymbol{q} \mid \boldsymbol{z}_i)$, where $\boldsymbol{q}$ is the input question.
The score is estimated by calculating the likelihood of generating the question $\boldsymbol{q}$ given the passage $\boldsymbol{z}_i$. The LLM computes the average log-likelihood of question tokens: \(\log p(\boldsymbol{q} \mid \boldsymbol{z}_i) = \frac{1}{|\boldsymbol{q}|} \sum_{t} \log p(q_t \mid \boldsymbol{q}_{<t}, \boldsymbol{z}_i; \Theta)\) where $\Theta$ are the parameters of the LLM. The generated question serves as a query, and we rank the passages based on how well the passage can produce the question. 
The second approach, called RankGPT~\cite{sun-etal-2023-chatgpt}, utilizes a permutation-based approach for re-ranking top-$K$ retrieved documents using LLMs like GPT-3.5 and GPT-4. Unlike traditional methods that evaluate documents independently, RankGPT optimizes the order of all retrieved documents by considering their relevance to a query.
RankGPT inputs a set of retrieved documents into an LLM, where each document is tagged with an identifier (e.g., $[1], [2]$, etc.). The LLM generates a ranking by outputting a permutation of these identifiers based on their relevance to the input query. This process directly produces a ranked list without relying on intermediate scoring, leveraging the LLM’s understanding of instructions for ranking.  

\subsection{Retrieval Augmented Generation Language model}
\label{sec:in-context-learning}

In-context learning enables LLMs to utilize external knowledge without altering their internal parameters. This approach enriches the context by incorporating both retrieved and generated documents directly into the model's input sequence. Given a sequence of tokens \(x_1, \ldots, x_n\), where \(x_1, \ldots, x_{i-1}\) represents the tokens preceding the current token \(x_i\), the goal is to predict \(x_i\). The standard formulation is:
 \(
p(x_1, \ldots, x_n) = \prod_{i=1}^n p_{\theta}(x_i | x_{<i}),
\)
where \(p_{\theta}(x_i | x_{<i})\) is the conditional probability of generating \(x_i\) based on its prefix \(x_{<i}\), with \(\theta\) representing the model’s parameters. We extend this by incorporating a set of retrieved documents \(\mathcal{R}(x_{<i})\) and a set of generated documents \(\mathcal{G}(q)\), where \(q\) is the query derived from the input prefix. The generation probability is:
 \(
p(x_1, \ldots, x_n) = \prod_{i=1}^n p_{\theta}(x_i | [x_{<i}; \mathcal{R}(x_{<i}); \mathcal{G}(q)]),
\)
where \([a; b; c]\) denotes the concatenation of sequences \(a\), \(b\), and \(c\). This setup allows the LLM to condition its output on both retrieved knowledge and newly generated content, providing a richer context for generation.

For each query \(q\) derived from the prefix tokens \(x_{<i}\), a set of top-$k$ relevant documents \(\{d_1, d_2, \ldots, d_k\}\) is retrieved, forming \(\mathcal{R}(q)\). Simultaneously, the model generates a set of contextual documents \(\mathcal{G}(q) = \{\mathbf{g}_1, \mathbf{g}_2, \ldots, \mathbf{g}_n\}\), where each \(\mathbf{g}_i\) is conditioned on \(q\). These documents serve as additional context to inform the answer generation. The combined influence of retrieved and generated documents allows the model to maximize the likelihood of generating the next token \(x_i\):
 \(
    d_{i^*}, g_{j^*} = \arg\max_{d \in \mathcal{R}(q), g \in \mathcal{G}(q)} p_{\theta}(x_i | [x_{<i}; d; g]),
\)
where \(d_{i^*}\) is the most relevant retrieved document and \(g_{j^*}\) is the most informative generated document for the given prefix and query. By balancing the factual accuracy of retrieved documents with the creative and contextual insights from generated documents, this hybrid approach enhances the model’s ability to address complex information needs. 
Building upon~\cite{ram-etal-2023-context}, which focuses on retrieval-augmented language models (RALMs), we extend the framework by incorporating generated documents into the retrieval process alongside retrieved passages. Unlike prior work, which primarily relies on external retrieval, we explore the trade-offs between retrieved and generated context within hybrid models, analyzing their impact on factual consistency and response diversity. Our analysis in Section~\ref{sec:exp-setup} reveals that while this hybrid retrieval-augmented generation approach improves recall in QA tasks, it introduces challenges related to retrieval redundancy, factual consistency, and ranking biases in hybrid models.

\section{Experimental Setup}\label{sec:exp-setup}

\begin{table*}[ht]
\centering
\resizebox{0.8\textwidth}{!}{
\begin{tabular}{l|cccccc|cccccc|cccccc}
\hline
\multirow{2}{*}{\textbf{Retriever}} & \multicolumn{6}{c|}{\textbf{NQ}} & \multicolumn{6}{c|}{\textbf{TQA}} & \multicolumn{6}{c}{\textbf{WEBQ}} \\
 & Top-1 & Top-5 & Top-10 & Top-20 & Top-50 & Top-100 & Top-1 & Top-5 & Top-10 & Top-20 & Top-50 & Top-100 & Top-1 & Top-5 & Top-10 & Top-20 & Top-50 & Top-100 \\
 \midrule
    \multicolumn{19
    }{c}{\textit{Unsupervised Retrievers}} \\
    \midrule

MSS         & 19.28 & 41.25 & 51.27 & 59.97 & 69.56 & 75.57 & 30.76 & 52.65 & 60.52 & 67.18 & 74.58 & 79.11 & 11.66 & 29.04 & 39.12 & 49.21 & 61.12 & 68.36 \\

BM25        & 22.11 & 43.77 & 54.46 & 62.94 & 72.74 & 78.25 & 46.30 & 66.28 & 71.74 & 76.41 & 80.56 & 83.15 & 18.90 & 41.83 & 52.17 & 62.40 & 71.70 & 75.49 \\
Contriever  & 22.16 & 47.29 & 58.73 & 67.87 & 76.01 & 80.55 & 34.16 & 59.49 & 68.00 & 73.91 & 79.84 & 82.94 & 19.98 & 43.45 & 56.40 & 65.70 & 74.85 & 80.12 \\
\midrule
\multicolumn{19}{c}{\textit{Supervised Retrievers}} \\
\midrule
DPR         & 48.67 & 68.78 & 74.54 & 79.20 & 83.71 & 85.71 & 57.47 & 72.40 & 76.50 & 79.77 & 82.97 & 85.10 & 44.83 & 65.01 & 70.62 & 74.61 & 78.69 & 81.64 \\

MSS-DPR     & \textbf{50.17} & \textbf{71.88} & \textbf{77.48} &\textbf{ 81.44} & \textbf{85.98} & \textbf{88.14} & 61.64 & 75.21 & 79.15 & 81.85 & 84.92 & 86.58 & 44.24 & 65.01 & 71.65 & 76.92 & 81.84 & 84.55 \\

\midrule
\multicolumn{19}{c}{\textit{Generator}} \\
\midrule
GenRead &  45.76 & 65.32 & 71.22 & - & - & - & \textbf{69.41} & \textbf{79.72} & \textbf{82.85} & - & - & - & \textbf{51.03} & \textbf{69.05} & \textbf{73.23} & - & - & - \\
\midrule

\end{tabular}}
\caption{Performance of different retrieval and generative models on Natural Questions (NQ), TriviaQA (TQA), and WebQuestions (WEBQ). 
}
\label{tab:retreivier_generator}
\end{table*}
\subsection{Datasets}\label{sec:datasets}
The evaluation is conducted for ODQA on Natural Questions (NQ)~\cite{kwiatkowski2019natural}, TriviaQA~\cite{joshi2017triviaqa} and WebQuestions (WebQ)~\cite{berant2013semantic}, following the same setup as in \cite{yu2022generate,izacard2020leveraging,lee2019latent}, while for Information Retrieval we use TREC~\cite{craswell2020overview} and  BEIR~\cite{thakur2021beir} and finally WikiText-103~\cite{merity2016pointer} for Language modelling.  

\textbf{ODQA datasets:} NQ is derived from real user queries made through Google Search, where answers are extracted as spans from Wikipedia articles. The dataset consists of 79,168 examples for training, 8,757 for development, and 3,610 for testing. TriviaQA, a dataset constructed from trivia and quiz-league websites, contains open-domain questions with well-defined answers. For ODQA, we use its unfiltered version, which includes 78,785 training examples, 8,837 development examples, and 11,313 test examples. WebQ consists of questions sourced via the Google Suggest API, with answers mapped to Freebase entities. It includes 3,417 training, 361 development, and 2,032 test examples.

\textbf{Evidence Passages:}
For ODQA, we used the preprocessed English Wikipedia dump from December 2018, as released by Karpukhin et al.~\cite{karpukhin2020dense}. Each Wikipedia article is split into non-overlapping passages of 100 words. This corpus contains over 21 million passages and serves as the evidence set from which relevant documents are retrieved for answering questions in QA tasks, we downloaded and used this corpus from Rankify, as described in \cite{abdallah2025rankify}.

\textbf{Information Retrieval (IR):} TREC~\cite{craswell2020overview} is a well-established benchmark dataset used in IR. For our evaluation, we utilize the test sets from the 2019 and 2020 TREC Deep Learning (DL) tracks: (i) TREC-DL19, which includes 43 queries, and (ii) TREC-DL20, comprising 54 queries. BEIR Benchmark~\cite{thakur2021beir}: is a heterogeneous benchmark covering 18 retrieval tasks, including fact-checking, question answering, and domain-specific retrieval (biomedical, scientific, etc.). 

\textbf{Language Modeling:} WikiText-103~\cite{merity2016pointer}: is a large-scale dataset of over 100 million tokens sourced from long, context-rich Wikipedia articles. This dataset is a standard for evaluating language modeling tasks.

All experiments and dataset processing were conducted using the Rankify\footnote{\url{https://github.com/DataScienceUIBK/Rankify}} framework, which provides a unified toolkit for retrieval, re-ranking, and retrieval-augmented generation~\cite{abdallah2025rankify}.

\subsection{\textbf{Retrieval and Generative Models}}
\textbf{Retrieval Models:}\label{sec:retrieval-models}
We used five retrieval models in our experiments: \textbf{BM25}, a sparse vector-based method; \textbf{DPR}, a dense dual-encoder model that maximizes similarity between questions and relevant passages; \textbf{MSS}, a dense retriever pre-trained on predicting masked spans like named entities; \textbf{MSS-DPR}, combining MSS pre-training with DPR’s fine-tuning for improved performance; and \textbf{Contriever}, an unsupervised dense retriever optimized for zero-shot performance through contrastive learning.

\begin{table}[t]
\centering
    \centering
    \resizebox{0.40\textwidth}{!}{  
    \begin{tabular}{l|cccccc|cccccc}
    \hline
    \multirow{2}{*}{\textbf{Retriever}} & \multicolumn{6}{c|}{\textbf{NQ}} & \multicolumn{6}{c}{\textbf{WEBQ}} \\
     & Top-1 & Top-5 & Top-10 & Top-20 & Top-50 & Top-100 & Top-1 & Top-5 & Top-10 & Top-20 & Top-50 & Top-100 \\
     \midrule
    \multicolumn{13}{c}{\textit{Generator+UPR}} \\
    \midrule
    GenRead &  44.76 &  64.18 & 71.22 & - & - & - & 51.72 & 67.13 & 73.23  & - & - & - \\
    \midrule
    \multicolumn{13}{c}{\textit{Retriever+UPR}} \\
    \midrule
    MSS        & 35.90 & 60.91 & 66.70 & 71.44 & 74.10 & 75.57 &  29.97 & 51.03 & 58.46 & 63.19 &  66.93 &  68.36 \\
    BM25        & 36.84 & 61.72  &68.45 & 72.63  & 76.68 & 78.25 & 33.12 & 56.45 & 63.73 &69.14 & 73.92 & 75.49 \\
    Contriever  &36.73 & 63.49 & 71.69 &76.32& 79.50& 80.55 & 33.96 & 59.94 & 67.18 & 73.08 & 78.20 & 80.12 \\
    DPR        &44.21 & 71.86 &  78.92 &82.16& 84.90 &  85.71 &  41.44 & 67.18 &72.93  &77.12 & 80.41 &   81.64 \\
    MSS-DPR    &  43.74 & 72.88 & 80.44 &84.71&87.26 & 88.14  &  39.96 & 65.40 & 73.08 & 78.30 & 82.63 &84.55  \\
    \midrule
    \multicolumn{13}{c}{\textit{Combined+UPR}} \\
    \midrule
    MSS        & 43.16 & 64.68 &  75.32 & 82.58 & 85.37 &86.29 & 48.87 & 66.49 &  72.83 & 78.89 &  81.20 &  82.33 \\
    BM25       & 43.46 & 64.52 &  75.73 &  82.96 & 86.09 & 87.09  & 48.28 & 66.78 & 73.33 & 78.94 &  82.38 & 83.76 \\
    Contriever & 43.27 & 64.99 & 76.34 &  83.93 &  87.15 & 87.78 & 47.93 & 67.67 & 73.97 & 79.48 &  83.56 & 84.99 \\
    DPR        & 44.07 &  65.90 &  78.17 &  86.68 & 89.39 & 90.11 &  49.46 &  68.06 &  74.90 &  81.89 &   85.19 &  86.27 \\
    MSS-DPR    &  43.96 &  65.60 &  77.87 &  87.51 &\textbf{ 90.42} &\textbf{ 91.27}  &  48.13 &  68.11 &  74.46 &  81.64 &  85.48 & \textbf{87.40} \\
    \midrule
    \multicolumn{13}{c}{\textit{Generator+RankGPT}} \\
    \midrule
    GenRead    &  50.97 &  64.74 & 71.22 & - & - & -  &  55.71 & 67.77 & 73.23 & - & - & - \\
    \midrule
    \multicolumn{13}{c}{\textit{Retriever+RankGPT}} \\
    \midrule
    MSS        &  43.52 & 63.19 & 68.34 & 70.28 & 73.85 & 75.57 &  35.58 & 53.30 & 58.76  &  61.91 & 65.8 & 68.36  \\
    BM25       &  48.98 & 66.76 & 70.86 & 73.71 & 76.40 & 78.25 &    42.27 &60.19   & 65.75 & 69.39 &  73.43  & 75.49 \\
    Contriever &  46.87 &  67.09 &71.58  &75.29  &  78.98 & 80.55 &   41.93&  63.24&   68.8&  73.23&77.12  &  80.12 \\
    DPR        &  50.47 & 75.24 & 80.00 & 82.71 & 84.88 & 85.71  &   48.28 &  68.85 &   74.26 &  77.31 & 79.82 & 81.64  \\
    MSS-DPR    & 54.88 & 75.35 & 81.47 & 84.88& 87.20 &88.14   &   49.56 &  69.83 &   75.15 &  79.28 & 82.43 & 84.55  \\
    \midrule
    \multicolumn{13}{c}{\textit{Combined+RankGPT}} \\
    \midrule
    MSS        & 53.60 & 68.37 & 76.95 & 81.88& 83.96 &86.12 &  54.77 & 68.6 & 73.72  & 76.43 & 79.53 & 82.04  \\
    BM25       & 53.46 & 68.48 & 76.98 & 82.22 &84.65  &  86.87 &  55.07 & 69.29 & 75.34  & 78.35 & 81.5 & 83.86  \\
    Contriever &  53.05 & 68.25 & 77.26 & 83.05 & 85.84 & 87.67&  55.86 & 68.90 & 74.51  & 78.74 &  82.38 &84.84  \\
    DPR        & 56.15 &  69.70 &  80.08 & 86.93 & 88.92 & 90.06 &  56.10 & \textbf{70.52} & \textbf{78.30}  & 82.33 & 84.25 & 85.93  \\
    MSS-DPR    & \textbf{56.79} & \textbf{70.55} &\textbf{80.58}  & \textbf{87.73} & 89.81& 91.16 &  \textbf{56.50} & 69.59 &  77.90  &  \textbf{82.53} & \textbf{85.63} &  87.30  \\
    \midrule
    \end{tabular}}
    \caption{Performance comparison of various retrieval and generation methods combined with UPR and RankGPT for reranking on NQ and WebQ datasets.}
    \label{tab:combined_reranking}
\end{table}
\textbf{Generative Models:}\label{sec:generative-models} For the generation-based retrieval, we employ GenRead~\cite{yu2022generate} a generative model designed for open-domain QA tasks, which first generates contextual documents based on the query and then predicts the final answer using those generated documents.

\subsection{Language Models}\label{sec:language-models}

We tested a range of LLMs in our experiments, focusing on both generation and reranking tasks: \textbf{GPT-2}~\cite{radford2019language}: A transformer-based autoregressive language model trained on WebText. We experimented with the small (110M), medium (345M), large (774M), and extra-large (1.5B) versions of GPT-2 to observe how model size impacts performance. \textbf{OPT}~\cite{zhang2022opt}: We experimented with various OPT models ranging from 125M to 13B parameters to analyze their performance across retrieval and generation tasks. \textbf{GPT-Neo}~\cite{gpt-neo}: An autoregressive language model trained on the Pile dataset~\cite{leo-etal-pile}. We evaluated its performance on WikiText-103 using both retrieval and generation configurations.
\begin{table*}[t]
\centering
\resizebox{0.7\textwidth}{!}{
\begin{tabular}{@{}l|c|ccc|ccc|ccc|ccc|ccc|ccc@{}}
\toprule
 \multirow{2}{*}{\textbf{Retriever} } & \multirow{2}{*}{\textbf{\# Mode}} & \multicolumn{3}{c|}{\textbf{LLama-3.3 8B}} & \multicolumn{3}{c|}{\textbf{LLama-3.1 8B}} & \multicolumn{3}{c|}{\textbf{Gemma-2-2b}} & \multicolumn{3}{c|}{\textbf{Gemma-2-9b}} & \multicolumn{3}{c|}{\textbf{Llama-2-13b-hf}} & \multicolumn{3}{c}{\textbf{Mistral-7B-v0.1}} \\

 & & \textbf{NQ} & \textbf{TriviaQA} & \textbf{WebQA} & \textbf{NQ} & \textbf{TriviaQA}& \textbf{WebQA} & \textbf{NQ} & \textbf{TriviaQA}& \textbf{WebQA} & \textbf{NQ} & \textbf{TriviaQA}& \textbf{WebQA} & \textbf{NQ} & \textbf{TriviaQA}& \textbf{WebQA} & \textbf{NQ} & \textbf{TriviaQA}& \textbf{WebQA} \\
\midrule
& G & 24.68 & 52.23 & 15.45 & 21.49 & 48.74 & 14.51 & 27.01 & 59.91 & 19.34 & 28.28 & 63.02 & 18.65 & 28.06 & 62.64 & 20.32 & 27.01 & 62.64 & 16.09 \\

\midrule
\multirow{6}{*}{BM25 } 
& R & 14.90 & 42.10 & 10.23 & 12.82 & 40.13 & 9.25 & 14.02 & 43.28 & 14.71 & 19.81 & 57.55 & 14.96 & 21.14 & 57.90 & 19.54 & 11.19 & 52.85 & 6.40 \\

& R+G & 25.51 & 53.29 & 15.05 & 22.29 & 49.69 & 13.97 & 28.39 & \textbf{59.89} & 19.29 & 28.45 & \textbf{63.50} & 19.05 & 26.62 & 61.35 & 19.00 & 25.68 & 60.45 & 15.65 \\

& G+R & 25.67 & 53.24 & 15.00 & 23.29 & 50.29 & 13.43 & 28.50 & 59.87 & 19.73 & 28.42 & 62.94 & 19.34 & 26.79 & \textbf{\underline{62.01} } & 19.24 & 23.71 & 58.56 & 13.44 \\

& UPR & 23.49 & \textbf{\underline{55.85} }& 17.27 & 24.24 & \textbf{55.22} & 17.67 & 26.23 & 58.71 & 19.78 & 23.41 & 58.74 & 15.94 & 27.59 & 61.60 & 20.37 & 25.18 & \textbf{\underline{59.64} } & \textbf{17.18} \\

& RankGPT & \textbf{28.45} & - & \textbf{19.73} & \textbf{26.65} & - & 18.55 & \textbf{30.36} & - & \textbf{21.11} & \textbf{30.75} & - & \textbf{\underline{21.06}} & \textbf{29.22} & - & \textbf{21.99} & \textbf{25.35} & - & 17.18 \\

\midrule
\multirow{6}{*}{MSS}
& R & 12.82 & 31.90 & 7.38 & 11.19 & 30.74 & 6.88 & 13.96 & 33.05 & 14.71 & 19.78 & 50.93 & 14.96 & 21.52 & 51.75 & 20.62 & 11.08 & 42.69 & 6.40 \\


& R+G & 25.54 & 51.39 & 14.96 & 21.57 & 47.75 & 14.46 & 28.48 & \textbf{58.63} & 19.29 & 28.56 & 62.53 & 18.80 & 26.81 & 58.64 & 19.05 & \textbf{\underline{25.90}} & 56.28 & 14.57 \\

& G+R & 25.31 & 51.81 & 15.20 & 22.68 & 48.75 & 13.43 & 28.42 & 58.47 & 19.78 & 28.31 &\textbf{ 61.78} & 18.55 & 27.48 & \textbf{59.56 }& 18.75 & 23.10 & 52.15 & 13.09 \\

& UPR & 23.35 & \textbf{53.78} & 17.08 & 24.43 & \textbf{54.97} & 16.70 & 26.20 & 58.15 & 19.73 & 23.10 & 57.50 & 15.85 & 27.29 & 59.23 & 20.72 & 23.77 & \textbf{57.24} & \textbf{\underline{17.42} }\\

& RankGPT & \textbf{28.17} & - & \textbf{19.05} & \textbf{25.84} & - & \textbf{17.37} & \textbf{29.17} & - & \textbf{19.93} & \textbf{29.97} & - & \textbf{19.64} & \textbf{27.56} & - & \textbf{20.77} & 23.77 & - & 16.98 \\

\midrule
\multirow{6}{*}{Contriever} 
& R & 15.29 & 36.25 & 10.67 & 13.24 & 35.29 & 9.35 & 13.96 & 33.05 & 14.71 & 19.78 & 50.93 & 14.96 & 20.47 & 42.69 & 19.98 & 11.08 & 42.69 & 6.40 \\


& R+G & 25.59 & 51.78 & 15.10 & 22.18 & 48.38 & 13.87 & 28.78 & \textbf{58.86} & 20.28 & 28.84 & \textbf{62.85} & 19.59 & 27.45 & 56.43 & 19.59 & \textbf{25.35} & 56.43 & 15.31 \\

& G+R & 25.70 & 52.07 & 15.50 & 23.13 & 48.91 & 13.87 & 28.75 & 58.64 & 20.13 & 28.37 & 61.97 & 19.09 & 27.17 & 52.66 & 19.88 & 23.02 & 54.39 & 13.44 \\

& UPR & 23.24 & \textbf{53.48} & 17.32 & 24.21 & \textbf{55.71} & \textbf{17.62} & 26.26 & 58.37 & 19.83 & 23.21 & 57.64 & 16.14 & 26.57 & \textbf{59.60} & \textbf{20.72} & 25.10 & \textbf{57.26} & \textbf{17.27} \\

& RankGPT & \textbf{30.55} & - & \textbf{19.78} & \textbf{28.86} & - & \textbf{17.62} & \textbf{32.11} & - & \textbf{20.67} & \textbf{32.44} & - & \textbf{19.88} & \textbf{30.39} & - & \textbf{20.72} & 25.10 & - & \textbf{17.27} \\

\midrule
\multirow{6}{*}{DPR} 
& R & 28.08 & 45.88 & 19.83 & 23.21 & 43.62 & 14.32 & 13.99 & 33.05 & 14.71 & 19.78 & 50.93 & 14.96 & 21.94 & 51.07 & 19.83 & 11.11 & 42.69 & 6.40 \\


& R+G & 28.94 & 54.41 & 24.62 & 24.62 & 50.61 & 14.32 & 30.25 &\textbf{ 60.16 }& 20.18 & 30.83 & \textbf{\underline{63.72}} & 19.93 & 28.81 & 58.00 & 20.32 & 27.70 & 58.00 & 16.44 \\

& G+R & 28.14 & \textbf{54.50} & \textbf{25.51} & 25.01 & 51.65 & 14.81 & 27.92 & 60.17 & 20.72 & 29.92 & 63.18 & 19.39 & 27.92 & 60.01 & 20.72 & 25.01 & 54.64 & 14.71 \\

& UPR & 23.60 & 53.41 & 18.06 & 24.74 & \textbf{\underline{56.07}} & \textbf{\underline{19.64}} & 26.51 & 58.71 & 19.78 & 23.38 & 57.85 & 16.04 & 27.45 & 60.05 & 20.62 & 25.25 & \textbf{57.53 }& 17.17 \\

& RankGPT & \textbf{31.74 }& - & 20.42 & \textbf{29.58} & - & \textbf{\underline{19.64}} & \textbf{\underline{34.16}} & - & \underline{\textbf{22.15} } & \underline{\textbf{34.04} }& - & \textbf{21.01} & \textbf{32.77} & - & \underline{\textbf{22.15} }& \textbf{25.26} & - &\textbf{ 17.18} \\

\midrule
\multirow{6}{*}{MSS+DPR}
& R & 28.17 & 47.69 & 23.68 & 23.68 & 45.72 & 13.92 & 13.96 & 33.05 & 14.71 & 19.78 & 50.93 & 14.96 & 21.47 & 51.35 & 19.83 & 11.08 & 42.69 & 6.40 \\


& R+G & 29.41 & \textbf{54.53} & 24.09 & 24.09 & 50.72 & 14.96 & 28.48 & \textbf{\underline{60.56} } & 19.29 & 28.45 & \textbf{62.53 }& 18.96 & 28.45 & 59.50 & 19.78 & 27.40 & 58.03 & \textbf{16.44} \\

& G+R & 28.61 & 54.48 & \textbf{\underline{25.54} }& 25.07 & 52.20 & 14.76 & 28.61 & 60.25 & 19.78 & 28.31 & 61.78 & 18.55 & 28.73 & 59.96 & 20.77 & 25.07 & 54.39 & 14.96 \\

& UPR & 23.10 & 53.65 & 16.68 & 25.24 & \textbf{55.52} & 18.98 & 23.10 & 58.64 & 16.68 & 25.24 & 57.47 & 16.09 & 23.10 & \textbf{60.00} & 19.98 & 25.23 & \textbf{57.47 }& 16.08 \\

& RankGPT & \textbf{\underline{31.94}} & - & 21.41 & \textbf{\underline{29.95}} & - & \textbf{18.99} & \textbf{32.51} & - & \textbf{21.41} & \textbf{32.13} & - & \textbf{18.99} & \textbf{\underline{32.61} }& - & \textbf{21.95} & \textbf{25.24} & - & 16.09 \\

\bottomrule
\end{tabular}
}
\caption{Zero-shot results of in-context learning on
the test set of NQ, TriviaQA, and WebQ measured by exact match.  
\textbf{Bold values} indicate the best performance within each retriever, while  
\underline{\textbf{underlined values}} represent the best overall performance. Please refer to the Appendix for additional evaluation metrics (e.g., Recall and F1-score) presented in Tables~\ref{tab:qa_LLama_3_3.1}, \ref{tab:qa_Gemma}, and \ref{tab:qa_Llama-2-13b}. }
\label{tab:em_comparison}
\end{table*}
\subsection{Reranking Methods}\label{sec:reranking-techniques}

We explored two reranking techniques to optimize the combination of retrieved and generated documents: \textbf{UPR}: Based on the T5 series~\cite{raffel2020exploring}, which consists of encoder-decoder transformers pre-trained on text denoising tasks. We used the T5-lm-adapt~\cite{lester2021power} and T0~\cite{sanh2022multitask} models for reranking. \textbf{RankGPT}: A reranking approach using GPT-3.5 and LLama 3 (70B) to evaluate the relevance of both retrieved and generated documents. These models dynamically rank documents to ensure the most relevant results are presented in QA tasks.

\begin{table}[ht]

    \centering
    \setlength\tabcolsep{5pt}
    \resizebox{0.40\textwidth}{!}{
    \begin{tabular}{l|ccc}
    \toprule
    \multirow{1}{*}{Models}& \multicolumn{1}{c}{NQ}  &  \multicolumn{1}{c}{TriviaQA} & \multicolumn{1}{c}{WebQ}   \\
    \midrule
    \multicolumn{4}{c}{\textbf{Retriever Only}} \\
    \midrule
    \textbf{REALM}~\cite{guu2020retrieval}&   40.4  &  - &  40.7    \\
    \textbf{DPR}~\cite{karpukhin2020dense}&   41.5   & 56.8 &  41.1  \\
    \textbf{RAG}~\cite{lewis2020retrieval}& 44.5   & 56.1 &45.2  \\
    \textbf{FiD-l}~\cite{izacard2020leveraging}&46.7   & 61.9 & 48.1 \\
    \textbf{FiD-xl}~\cite{izacard2020leveraging}&50.1   & 66.3 & 50.8 \\
    \textbf{FiD-xl}~\cite{izacard2020leveraging} &45.0  &70.1 &53.6   \\
    \textbf{FiD}~\cite{izacard2020leveraging}&51.4  &67.6 &50.5    \\
    \textbf{EMDR}~\cite{singh2021end}& 52.5  & 71.4 & 48.7  \\
    \textbf{RFiD-large}~\cite{wang2023rfid} &  54.3   & 72.6 & -  \\
    \textbf{DensePhrases}~\cite{lee2020learning}&14.5  &34.4 & 17.3   \\
    \textbf{DensePhrases}~\cite{lee2021learning}& 41.3  &53.5 & 41.5  \\
    \midrule
    \multicolumn{4}{c}{\textbf{Generator}} \\
    \midrule
    \textbf{GenRead (FiD-l)}~\cite{yu2022generate}  & 40.3 & 67.8 & 51.5   \\
    \textbf{GenRead (FiD-l)}~\cite{yu2022generate}   & 43.5& 70.2 & 53.3   \\
    \textbf{GenRead (FiD-xl)}~\cite{yu2022generate}   & 42.6& 69.6 & 52.6  \\
    \textbf{GenRead (FiD-xl)}~\cite{yu2022generate}  & 45.6  & 71.6 &\textbf{ 54.4 } \\
    \midrule
    \multicolumn{4}{c}{\textbf{Generator and Retriever}} \\
    \midrule
    Combine Document &  \textbf{57.4}  &  \textbf{75.7}  & 53.6   \\
    \bottomrule
    \end{tabular}}
    \caption{Exact match (EM) performance of Llama 2-7B trained on retrieved and generated documents compared to baseline models across NQ, TriviaQA, and WebQ datasets using DPR and Generated Documents.}
    \label{tab:odqaresult}
\end{table}
\section{Open-Domain QA Results}
\subsection{Retrieval and Generation Results:}

In this section, we evaluate the performance of various retrieval-based models and a generative model, GenRead, on three open-domain QA datasets: \texttt{NQ}, \texttt{TriviaQA}, and \texttt{WebQ}. We compare unsupervised retrievers (BM25, MSS, and Contriever), supervised retrievers (DPR and MSS-DPR), and the GenRead generative model to understand their effectiveness in different retrieval settings. Table \ref{tab:retreivier_generator} presents the results for Top-1, Top-5, Top-10, Top-20, Top-50, and Top-100 retrieval accuracies across the three datasets. For the unsupervised retrieval models, Contriever achieves the highest Top-1 accuracy on NQ, outperforming BM25 and MSS by capturing deeper semantic relationships between queries and passages, consistent with findings by~\cite{izacard2021unsupervised}. However, the supervised retrievers show a clear advantage, with MSS-DPR achieving the highest Top-1 accuracy of 50.17\% on NQ, demonstrating the impact of training on specific QA datasets like NQ~\cite{sachan2021end}.
\begin{table}[!t]
\centering
\small
\setlength\tabcolsep{2pt}
\begin{adjustbox}{width=0.5\textwidth,center}
\begin{tabular}{l cc | cccccccc | c }

\toprule
\textbf{Method} & DL19 & DL20 & Covid &  NFCorpus &  Touche &  DBPedia & SciFact &  Signal & News &  Robust04 & BEIR (Avg) \\

\midrule

BM25
& 50.58 & 47.96 & 59.47 & 30.75 & 44.22 & 31.80 & 67.89 & 33.05 & 39.52 & 40.70 & 43.42\\

\midrule
\textbf{Supervised} \\
\midrule

monoBERT (340M)
& 70.50 & 67.28 & 70.01 & 36.88 & 31.75 & 41.87 & 71.36 & 31.44 & 44.62 & 49.35 & 47.16
\\

monoT5 (220M)
& 71.48 & 66.99 & 78.34 & 37.38 & 30.82 & 42.42 & 73.40 & 31.67 & 46.83 & 51.72 & 49.07
\\

monoT5 (3B)
& 71.83 & 68.89 & 80.71 & \textbf{\underline{38.97} } & \textbf{32.41} & \textbf{44.45} &  \underline{\textbf{76.57}} & 32.55 & 48.49 & \textbf{56.71} & \textbf{51.36}
\\

TART-Rerank (FLAN-T5-XL)
& -&- &75.10  &36.03 & 27.46 & 42.53 & 74.84 & 25.84 & 40.01 & 50.75 & -
\\

Cohere Rerank-v2
& \textbf{\underline{73.22} } & \textbf{67.08} & 81.81 & 36.36 & 32.51 & 42.51 & 74.44 & 29.60 & 47.59 & 50.78 & 49.45
\\

 Cohere Embedding-large&-&-
 & 80.10 & 34.70 & 27.60 & 37.20 & 72.10 & 30.60 & 46.10 & 48.90 & 47.16
 \\

 OpenAI Embedding-ada&-&-
 & \textbf{81.30} & 35.80 & 28.00 & 40.20 & 73.60 & \textbf{32.90} & \textbf{49.50} & 50.90 & 49.02
\\

\midrule
\textbf{Unsupervised} \\
\midrule

UPR (FLAN-T5-XL)
& 53.85 & 56.02 & 68.11 & 35.04 & 19.69 & 30.91 & 72.69 & 31.91 & 43.11 & 42.43 & 42.99
\\




 Promptagator++ (zero-shot)$^\dagger$
 & - & - & 76.0 & 36.0 & 27.8 & 41.3 & 73.6 & - & - & - & -
 \\

Promptagator++ (few-shot)
& - & - & 76.2 & 37.0 & 38.1 & 43.4 & 73.1 & - & - & - & -
\\

BM25+RankGPT & 65.80 & 62.91&  76.67&  \textbf{35.62} &\textbf{\underline{36.10} }& 44.47 & 70.43 & 32.12 & 48.85 &50.62 & 49.37\\

BM25+GenRead+RankGPT & \textbf{72.21} & \textbf{\underline{68.93}} & \textbf{\underline{82.81} } &  38.59 &  35.82 & \textbf{\underline{45.64}} &  \textbf{75.46} &  \textbf{\underline{33.64} }& \textbf{\underline{50.35} }& \textbf{\underline{58.48} }& \textbf{\underline{52.59} }\\


\bottomrule
\end{tabular}
\end{adjustbox}

\caption{Results (nDCG@10) on TREC and BEIR. Best performing unsupervised and overall system(s) are marked bold. For generating documents we used LLama-3.1 70B.
}
\label{table:beir_benmark}

\end{table}

The generative model, GenRead, achieves competitive performance, particularly in TriviaQA, where it outperforms all retrievers with a Top-1 accuracy of 69.41\%. This indicates the model's ability to generate contextually relevant passages even when traditional retrieval may fall short. However, generating more passages like retrieving documents (such as Top-50 or Top-100) increases the computational costs associated with generating documents and the risk of repetitive content~\cite{ganguli2022predictability}. Generative models require significant resources to create contextually relevant documents, which becomes increasingly demanding as the number of generated documents increases. Additionally, there is a tendency for generative models to produce similar or redundant outputs when tasked with generating numerous responses for a single query~\cite{ganguli2022predictability}.

\subsection{ Re-ranking Results}

In this section, we evaluate the impact of combining retrieval and generation through reranking methods, specifically using UPR and RankGPT, to refine document selection for open-domain tasks. Table \ref{tab:combined_reranking} illustrates the performance of different retrieval models paired with UPR and RankGPT across NQ and WebQ datasets. UPR enhances the precision of document ranking, as evidenced by DPR's improvement in Top-10 accuracy from 74.54\% to 80.44\% on the NQ dataset when combined with UPR. RankGPT, which leverages the semantic capabilities of large language models for reranking, achieves further gains, particularly for hybrid methods. For example, MSS-DPR with RankGPT achieves a Top-10 accuracy of 81.47\% on NQ, compared to 77.87\% without reranking. Combining retrieval and generative outputs, especially with methods like \texttt{Combined+RankGPT}, provides a balance between broad retrieval coverage and specific contextual information from LLMs. This is demonstrated by the MSS-DPR method achieving a Top-100 accuracy of 91.16\% on NQ when paired with RankGPT. Additionally, on the TriviaQA dataset as shown in Figure \ref{fig:tqa_retreiver_combined_reranking} (see Appendix~\ref{app:trivialqa}), the UPR method's performance comparison reveals that the GenRead+UPR retriever reaches the highest Top-1 accuracy at 69.74\%, while DPR+GenRead+UPR and MSS-DPR+Gen+UPR follow closely with Top-1 accuracies of 67.50\% and 67.30\%, respectively. Hybrid methods like BM25+Gen+UPR and DPR+Gen+UPR excel in Top-10 accuracy, achieving 84.41\% and 85.03\%, respectively, showing the benefit of combining generative context with retrieval outputs. However, we note that RankGPT's reranking process is computationally expensive, costing over $1,500$ for evaluations across NQ and WebQA, which limited its application to only UPR for the TriviaQA dataset in our experiments.


\begin{figure*}[!t]
    \centering
    \includegraphics[width=0.8\linewidth]{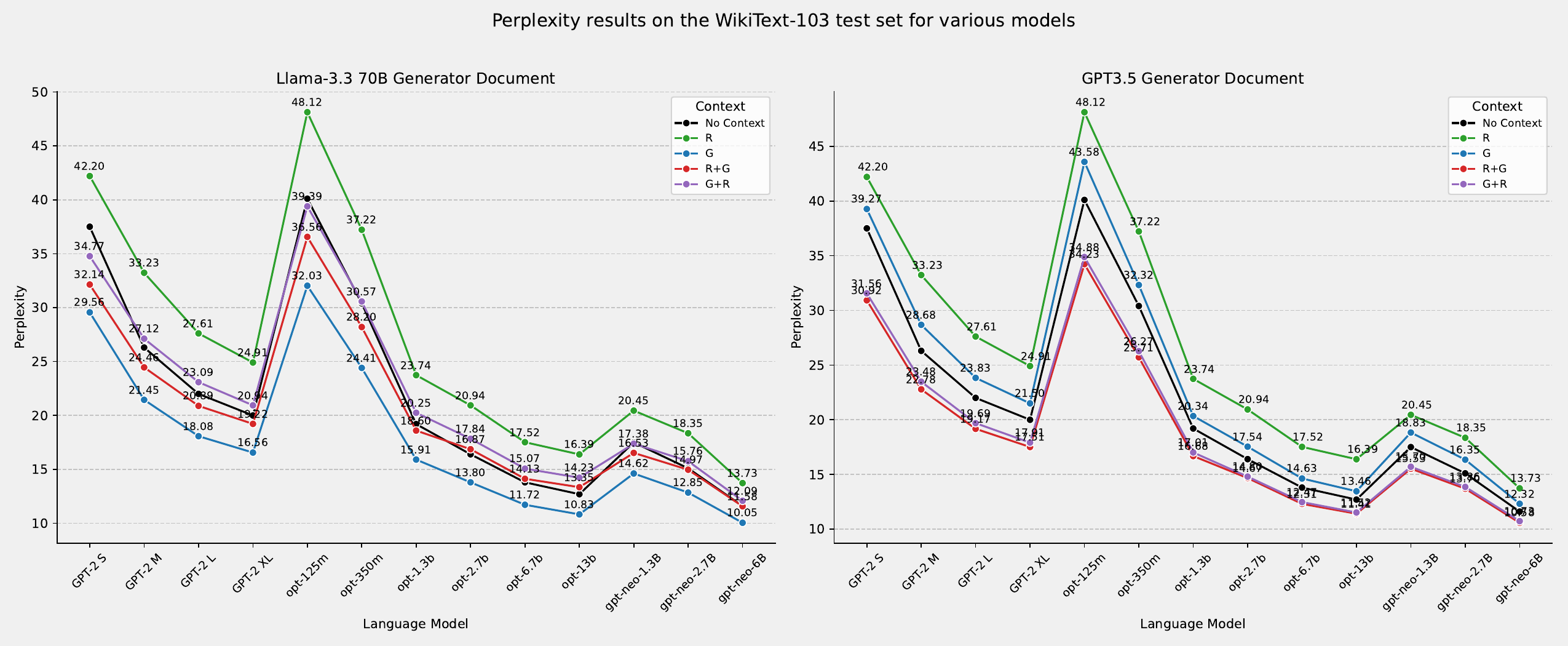}
    \caption{Perplexity Comparison for Language Modeling with Retrieval, Generation, and Hybrid Context Strategies using Different Generator Documents. The figure is divided into two subplots, each representing a different document generator used for providing context: (a) Llama-3 70B Generator Document and (b) GPT3.5 Generator Document. In both subplots, language model perplexity is evaluated under several context strategies: 'No Context' (baseline), 'R' (Retrieval-only using BM25 from Wikipedia), 'G' (Generation-only, context from a generated document), 'R+G' (Retrieval followed by Generation), and 'G+R' (Generation followed by Retrieval).  
    }
    \label{fig:Perplexity-language-model}
\end{figure*}

\subsection{In-Context for Open-Domain QA}

In this section ,we evaluate a Retrieval-Augmented Generation (RAG) setting, where retrieval-based models (BM25, MSS-DPR) are combined with generative models (e.g., InstructGPT, GenRead). The retrieved documents serve as input to a large language model, which generates the final response. 
Similar to prior works~\cite{karpukhin-etal-2020-dense,izacard2020leveraging}, our setup involves retrieval (R), generation (G), and hybrid models (R+G, G+R). Furthermore, we implement reranking techniques such as UPR and RankGPT to refine the document selection process. Our reader model (LLama-3.3 8B) gets the question along with its corresponding retrieved documents and returns the answer. Reader models are simply a frozen large LM (not pre-trained, fine-tuned).  Table~\ref{tab:em_comparison}, shows that hybrid models (R+G, G+R) when combined with reranking approaches such as RankGPT offer a more balanced approach across various datasets. For example, \texttt{BM25 R+G} achieves a score of 25.51 on NQ, in comparison with both retrieval-only (14.90) and generation-only (24.68) models in the \texttt{LLama V3.3 8B} setup. In particular, RankGPT consistently enhances performance, with BM25+RankGPT achieving 28.45 on NQ and 19.73 on WebQA, which highlights the effectiveness of reranking in refining document selection for question answering.

\vspace{-2mm}
\subsection{Supervised open-domain QA}

In this section, we evaluate the Fusion-in-Decoder (FiD) model~\cite{izacard2020leveraging} using LLama-2 7B. We integrate retrieved documents from DPR with generated content, leveraging both retrieval and generative for enhanced question-answering EM.  We compare Finetuned LLama-2 7B with other Retrieve-Read models, including DPR~\cite{karpukhin2020dense}, REALM~\cite{guu2020retrieval}, RAG~\cite{lewis2020retrieval}, and  GENREAD~\cite{yu2022generate}. Table \ref{tab:odqaresult} presents the EM for Llama 2-7B trained on retrieved and generated documents across the benchmarks. As seen, DPR combined with the generative model achieves competitive results, with an EM score of 57.4 on the NQ test set, 75.7 on TriviaQA, and 53.6 on WebQ. This performance is compared with baseline models such as REALM, and the FiD variants, showing improvements in most cases. For instance, the Trained LLama Model on generated and retrieved outperforms FiD-xl (50.1 EM on NQ) with a 7.3\% increase when using DPR.

\vspace{-4mm}

\section{Language Modeling Performance}\label{sec:5}

In this section, we explore the performance of retrieval (R), generation (G), and hybrid models (R+G and G+R) for language modeling on their impact on perplexity, a widely used evaluation metric for language models. The models are evaluated on the \texttt{WikiText-103} test set using \texttt{GPT2}, \texttt{OPT} and \texttt{GPT-Neo}, as presented in Figure~\ref{fig:Perplexity-language-model} (see Table~\ref{tab:language_model_retrieval_results} in the Appendix). We aim to analyse how retrieving documents from Wikipedia affects perplexity and can generate documents that can help also like in the ODQA task. Table \ref{tab:language_model_retrieval_results} shows that using GPT-2 with BM25 retrieval achieves a perplexity of 29.56, outperforming the generation-based (Llama-3.3 70, GPT3.5) model, which yields a perplexity of 42.20 and 39.27. The perplexity of retrieval models is $\text{Perplexity} = \exp\left(- \frac{1}{N} \sum_{i=1}^{N} \log p_\theta(w_i | \text{context}) \right)$, where $N$ represents the total number of words and $p_\theta(w_i | \text{context})$,
where $w_i$ is the $i$-th word in the sequence, and $p_\theta(w_i | \text{context})$ represents the probability assigned by the model $\theta$ given the retrieved context.  Hybrid models that combine retrieval and generation in two configurations do not outperform retrieval-only models as seen in Table \ref{tab:language_model_retrieval_results}. For instance, in LLama-3.3 70B, the R+G setup yields a perplexity of 32.14, compared to the retrieval-only model's 24.91. 

\section{Information Retrieval Performance}

Finally, we present an evaluation of the \texttt{BM25+GEN+RankGPT} method against state-of-the-art supervised and unsupervised models for information retrieval on the TREC and BEIR benchmarks. We focus on nDCG@10 scores across BEIR datasets. The supervised baselines include monoBERT~\cite{nogueira2019passage}, monoT5~\cite{nogueira2020document}, and Cohere Rerank. The unsupervised baselines include UPR~\cite{sachan2022improving}, and Promptagator++~\cite{dai2022promptagator}. Table \ref{table:beir_benmark} presents that the \texttt{BM25+GEN+RankGPT} consistently outperforms BM25 across all benchmarks, achieving the highest nDCG@10 scores on BEIR and TREC datasets. For example, on Robust04 dataset, \texttt{BM25+GEN+RankGPT} achieves an nDCG@10 score of 58.48, compared to 43.42 for BM25. Similarly, on SciFact, the hybrid model reaches 45.64, outperforming both supervised and unsupervised baselines like monoT5 (44.45) and UPR (30.91).

\section{Conclusion}
This study compares retrieval-based, generation-based, and hybrid models across QA, reranking, information retrieval, and language modeling. Retrieval models like BM25 and DPR excel in factual accuracy, while generative models provide contextual richness but struggle with consistency. Hybrid models effectively balance retrieval and generation, enhancing QA and IR performance. However, in language modeling, hybrid and generative approaches do not consistently outperform retrieval-based methods, underscoring the importance of retrieval for factual accuracy. 



\section*{Limitations} While our study demonstrates promising results in open-domain question answering (ODQA), document reranking, and retrieval-augmented language modeling, several limitations warrant further attention:

\begin{enumerate} \item The computational complexity of hybrid models, which combine retrieval and generation, increases with both the size of the corpus and the length of documents. This can lead to slower processing times, especially for large-scale datasets. \item The effectiveness of dense retrievers like DPR is highly dependent on the quality and diversity of the corpus used for training. Poorly representative datasets may lead to reduced performance in real-world applications. \item While hybrid models show significant improvements in document reranking, they are sensitive to the interplay between the retrieval and generation components. Inconsistent alignment between these components could lead to suboptimal performance in certain scenarios. 
\item Our evaluation is primarily limited to standard benchmarks, such as NQ and BEIR, which may not fully capture the diverse nature of real-world knowledge-intensive tasks. Besides other types of questions and retrieval tasks, the analysis should be extended to domain-specific scenarios, especially ones with low tolerance for errors and hallucinations like Medical \cite{kim-etal-2024-medexqa} or Legal QA \cite{abdallah2023exploring}. \end{enumerate}

\section*{Ethical Considerations and Licensing}

Our research utilizes the GPT models, which is available under the OpenAI License and  Apache-2.0 license, and the Llama model, distributed under the Llama 3 Community License Agreement provided by Meta. We ensure all use cases are compliant with these licenses. Additionally, the datasets employed are sourced from repositories permitting academic use. We are releasing the artifacts developed during our study under the MIT license to facilitate ease of use and adaptations by the research community. We have ensured that all data handling, model training, and dissemination of results are conducted in accordance with ethical guidelines and legal stipulations associated with each used artifact.
\bibliography{custom}
\appendix

\section{Appendix}

This appendix provides detailed insights into the retrieval and reranking results discussed in the main paper. We present performance evaluations of different retrieval methods, including BM25, DPR, MSS, and hybrid approaches (R+G, G+R) across multiple models. The results demonstrate the impact of reranking and retrieval-augmented generation (RAG) techniques on various question-answering benchmarks.

\subsection{Retrieval Performance on TriviaQA}
\label{app:trivialqa}

Retrieval plays a fundamental role in the TriviaQA dataset, where models must extract relevant information from large document collections to answer trivia-based, open-domain questions. This section provides a detailed comparison of various retrieval methods, including sparse retrievers like BM25, dense retrievers such as DPR, and generator models. The retrieval effectiveness of these models is measured using Top-1, Top-5, and Top-10 accuracy, which represent the percentage of cases in which a correct document appears within the top-k retrieved results.

The results show that DPR achieves the highest Top-1 accuracy at 75.4\%, significantly outperforming BM25, which achieves only 54.0\%. This indicates that sparse retrieval methods struggle with the complexity of trivia-style questions, while dense retrieval models that leverage learned representations of queries and documents exhibit superior retrieval effectiveness. MSS-DPR follows closely, with a Top-1 accuracy of 73.5\%, suggesting that additional pretraining techniques further enhance retrieval performance. Generative augmentation also proves valuable, as GenRead achieves a Top-1 accuracy of 69.7\%, surpassing BM25 and approaching the effectiveness of dense retrievers.

The advantages of generator approaches become more evident in the Top-5 and Top-10 accuracy metrics. MSS-DPR+Gen leads with an 85.0\% Top-5 accuracy, followed closely by DPR+Gen at 84.4\%, indicating that the combination of retrieval and generation improves ranking effectiveness. BM25+Gen also sees significant improvements, achieving 84.4\% in Top-5 accuracy, compared to BM25 alone at 73.6\%. In the Top-10 retrieval setting, hybrid models consistently outperform retrieval-only methods, with DPR+Gen reaching 85.2\% and MSS-DPR+Gen achieving 85.0\%. These findings confirm that hybrid approaches, which integrate retrieval with generative document expansion, provide more robust and reliable retrieval for complex QA tasks.

\begin{figure*}[t!]
    \centering
    \begin{minipage}{0.8\textwidth}
        \centering
        \includegraphics[width=\textwidth]{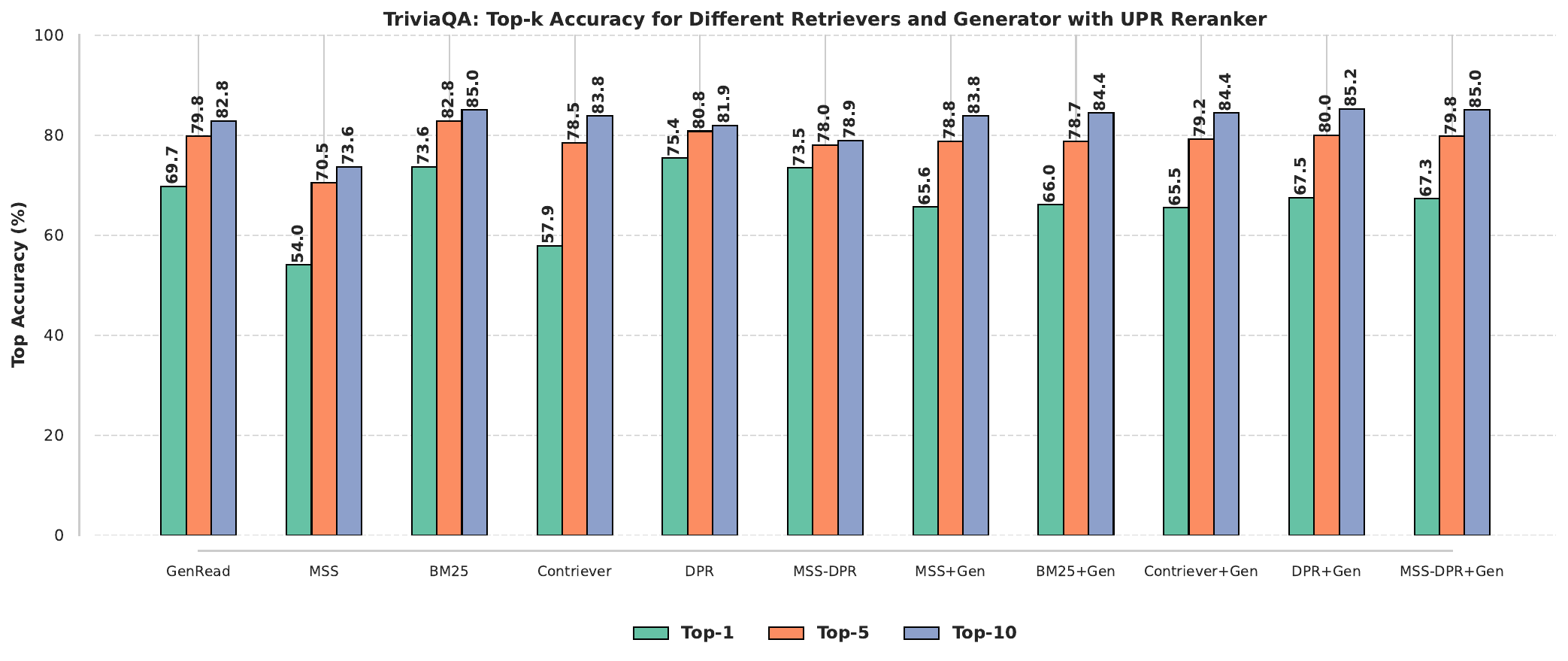}
        \caption{Comparison of retrieval methods using the UPR approach on the TriviaQA dataset, highlighting Top-1, Top-5, and Top-10 accuracy (Gen refers to GenRead method). The results showcase the impact of different retrieval and reranking strategies.
        }
        \label{fig:tqa_retreiver_combined_reranking}
    \end{minipage}\hfill
\end{figure*}
\subsection{Language Model Perplexity Evaluation}
This section evaluates the impact of retrieval on language modeling performance, measured using perplexity (PPL). Lower perplexity values indicate a model's ability to generate more fluent and contextually appropriate text, making it a key metric for evaluating generative language models. We compare various configurations, including retrieval-based (R), generative (G), and hybrid retrieval-generation approaches (R+G, G+R), across different model sizes, including GPT-2, OPT, and GPT-Neo, on the WikiText-103 benchmark.

The results indicate that retrieval (R) consistently improves language model performance, leading to substantial reductions in perplexity. For instance, GPT-2 Small achieves a perplexity of 37.5 without any augmentation, which drops to 29.56 with BM25-based retrieval, representing a 21.2\% improvement. Similarly, GPT-2 XL sees its perplexity reduced from 20.0 to 16.56 when retrieval is applied, highlighting the benefits of factual grounding in reducing language modeling uncertainty. Larger models such as OPT-13B achieve the lowest perplexity, with a reduction from 12.7 to 10.83 when retrieval is used.

Generative (G), in contrast, does not always lead to improvements in perplexity. In several cases, generating context without retrieval increases perplexity, as seen with GPT-2 XL, where perplexity rises from 20.0 to 24.91 when using generation alone. This suggests that generative models may introduce hallucinations, making their predictions less certain and increasing the likelihood of generating inaccurate text. The same trend is observed in smaller models, such as OPT-125M, where perplexity worsens from 40.1 to 48.12 with generation.

Hybrid approaches that combine retrieval and generation yield mixed results. The retrieval-first strategy (R+G) consistently outperforms the generation-first approach (G+R), as seen in GPT-2 XL, where R+G achieves a perplexity of 19.22 compared to 20.94 for G+R. The results confirm that retrieval should precede generation for maximum benefit, ensuring that the generative model conditions its output on factually accurate retrieved information. 

\begin{table}[t]
\centering
\setlength\tabcolsep{5pt}
\begin{adjustbox}{width=0.5\textwidth,center}
\begin{tabular}{@{}l|c|c|ccc|ccc@{}}
\toprule
\multirow{2.6}{0pt}{\textbf{Model}} & \multirow{2}{*}{No Context} & &\multicolumn{3}{c|}{\textbf{LLama-3.3 70B}  } & \multicolumn{3}{c}{\textbf{GPT3.5} }\\
\cline{4-9}
&  & R & G & R+G  & G+R  & G & R+G  & G+R\\
\midrule
\textbf{GPT-2 S} &  37.5& 29.56 & 42.20 & 32.14& 34.77   & 39.27 &30.92 & 31.56\\
\midrule
\textbf{GPT-2 M} &  26.3 & 21.45 &  33.23 & 24.46 & 27.12 & 28.68 & 22.78 & 23.48\\
\midrule
\textbf{GPT-2 L}  & 22.0 & 18.08 & 27.61 & 20.89  & 23.09 & 23.83 & 19.17 &  19.69\\
\midrule
\textbf{GPT-2 XL}  &   20.0&   16.56 &  24.91 & 19.22 & 20.94 & 21.50 &17.51 & 17.91\\
\midrule
\textbf{opt-125m}  &   40.1&  32.03  &  48.12 & 36.56 & 39.39 & 43.58 & 34.23 & 34.88\\
\midrule
\textbf{opt-350m}  &  30.4 &  24.41  & 37.22  &  28.20 & 30.57 & 32.32  & 25.71 &  26.27 \\
\midrule
\textbf{opt-1.3b}  &   19.2&  15.91  &  23.74 &  18.60 & 20.25 & 20.34& 16.68 &17.01 \\
\midrule
\textbf{opt-2.7b}  &   16.4& 13.80   &  20.94 & 16.87 &  17.84 & 17.54 &  14.67 & 14.80\\
\midrule
\textbf{opt-6.7b}  &   13.8&  11.72  &  17.52 &  14.13 & 15.07 & 14.63 & 12.31 &  12.47 \\
\midrule
\textbf{opt-13b}  &   12.7&  10.83  & 16.39  & 13.35 &  14.23 &13.46 & 11.41 & 11.52 \\
\midrule
\textbf{gpt-neo-1.3B}    & 17.5 &  14.62  &  20.45 & 16.53 & 17.38 &  18.83 & 15.53 & 15.70 \\
\midrule
\textbf{gpt-neo-2.7B}    & 15.1 &  12.85  & 18.35  & 14.97 &  15.76 & 16.35 & 13.70 & 13.86\\
\midrule
\textbf{gpt-neo-6B}    & 11.6 &  10.05  &  13.73  & 11.58 &  12.09 & 12.32 &  10.58 &  10.73\\

\bottomrule
\end{tabular}
\end{adjustbox}
\caption{
Perplexity results on the WikiText-103 test set for various models (GPT-2, OPT, GPT-Neo) using retrieval (R), generation (G), and hybrid approaches (R+G and G+R). Llama-3 70B and GPT3.5 used for generated Documents.
}
\label{tab:language_model_retrieval_results}
\end{table}

\subsection{QA performance comparison on multiple benchmarks}

This section presents a detailed zero-shot evaluation of various retrieval and reranking methods within a retrieval-augmented generation (RAG) framework. The performance of these methods is assessed across three widely used question-answering benchmarks: Natural Questions (NQ), TriviaQA, and WebQuestions (WebQ). The study compares different retriever and reranker models, including BM25, Multi-Stage Search (MSS), Contriever, Dense Passage Retrieval (DPR), and hybrid approaches such as retrieval + generation (R+G) and generation + retrieval (G+R). The results are reported for multiple state-of-the-art large language models, including LLaMA-3 8B, LLaMA-3.1 8B, Gemma-2 (2B and 9B), LLaMA-2-13B, and Mistral-7B-v0.1.

\subsubsection{Retrieval and RAG Performance on LLaMA-3 8B and LLaMA-3.1 8B}
The performance results for LLaMA-3 8B and LLaMA-3.1 8B across different retrieval approaches are presented in Table \ref{tab:qa_LLama_3_3.1}. The models are evaluated using three primary metrics: Exact Match (EM), Recall, and Consistency (Con). The EM score measures the percentage of responses that exactly match the ground truth, Recall represents the proportion of relevant documents retrieved, and Consistency assesses the stability of the model’s generated answers across multiple retrieval settings.

For LLaMA-3 8B, the DPR+G+R hybrid method achieves the highest EM score on TriviaQA (54.50\%) and NQ (28.14\%), while for WebQ, the BM25+G and DPR+G methods perform comparably with 15.45\% and 15.50\% EM, respectively. In contrast, traditional BM25 retrieval alone exhibits significantly lower performance, achieving only 14.90\% EM on NQ, 42.10\% on TriviaQA, and 10.23\% on WebQ. The R+G (Retriever first, then Generator) approach consistently outperforms G+R, with DPR+R+G reaching 28.94\% EM on NQ and 37.31\% on TriviaQA. This demonstrates that conditioning retrieval before generation is a superior strategy for answer synthesis.

For LLaMA-3.1 8B, a general performance improvement is observed over LLaMA-3 8B. The DPR+R+G method improves EM on NQ to 30.83\% and TriviaQA to 37.89\%, indicating that a refined retriever-generator interaction further enhances retrieval-augmented generation. The MSS+DPR approach achieves a strong balance between retrieval recall and consistency, with Recall reaching 49.23\% on NQ and 73.67\% on TriviaQA, while maintaining high consistency.

\subsubsection{Performance of RAG on Gemma-2 (2B and 9B) Models}
The retrieval and reranking evaluation on Gemma-2-2B and Gemma-2-9B is presented in Table \ref{tab:qa_Gemma}. The larger Gemma-2-9B model significantly outperforms its 2B counterpart across all benchmarks, demonstrating the benefits of model scaling in retrieval-augmented question answering.

On Gemma-2-2B, DPR+R+G achieves 30.25\% EM on NQ and 37.73\% on TriviaQA, while the traditional BM25 retriever falls behind, achieving 14.02\% EM on NQ and 43.28\% on TriviaQA. The generative augmentation (G) alone leads to better recall but performs worse in terms of EM than retrieval-augmented methods, with 46.85\% Recall on NQ and 72.99\% on TriviaQA.

The Gemma-2-9B model exhibits substantial improvements, with DPR+R+G achieving 30.83\% EM on NQ and 37.89\% on TriviaQA, mirroring the performance trends observed with LLaMA-3.1 8B. Interestingly, BM25+G surpasses the retrieval-only (BM25 R) approach by achieving 63.02\% Recall on TriviaQA, reinforcing that generative augmentation benefits sparse retrieval methods. However, the best-performing approach remains MSS+DPR+R+G, which achieves a Reciprocal Rank of 49.23\% on NQ and 73.67\% on TriviaQA, emphasizing the importance of hybrid search.

\subsubsection{Retrieval-Augmented QA with LLaMA-2-13B and Mistral-7B-v0.1}
Table \ref{tab:qa_Llama-2-13b} presents the results for LLaMA-2-13B and Mistral-7B-v0.1, two widely used open-source models. LLaMA-2-13B achieves better retrieval performance than Mistral-7B, especially when combining dense retrieval with reranking.

For LLaMA-2-13B, DPR+R+G consistently achieves the highest scores, with 30.91\% EM on NQ and 38.12\% on TriviaQA, surpassing the BM25+R baseline (21.14\% EM on NQ and 57.90\% on TriviaQA). However, BM25+G shows better recall (73.63\% on TriviaQA) compared to dense retrieval methods, supporting the argument that generative augmentation enhances sparse retrieval. The hybrid approach (DPR+G+R) further improves retrieval, attaining EM scores of 30.66\% on NQ and 37.98\% on TriviaQA, slightly trailing behind DPR+R+G but still outperforming retrieval-only baselines.

For Mistral-7B-v0.1, the results indicate that BM25 alone is highly ineffective, achieving only 11.19\% EM on NQ and 52.85\% on TriviaQA. The DPR+G+R hybrid model achieves 25.07\% EM on NQ and 28.92\% on TriviaQA, demonstrating that retrieval-first approaches remain more effective than generation-first pipelines. Interestingly, UPR (Unsupervised Passage Reranking) achieves 25.24\% EM on NQ and 31.25\% on TriviaQA, proving to be a strong alternative to traditional DPR reranking.

\begin{table*}[!ht]
\centering
\resizebox{1.0\textwidth}{!}{  
\setlength\tabcolsep{3pt}
\begin{tabular}{@{}l|c | ccc| ccc | ccc |ccc | ccc | ccc @{}}
\toprule
 & & \multicolumn{9}{c|}{\textbf{LLama V3.3 8B}} & \multicolumn{9}{c}{\textbf{LLama V3.1 8B}}\\

\multirow{3}{*}{\textbf{Retriever} }& \multirow{3}{*}{\textbf{\# Doc}} & \multicolumn{3}{c}{\textbf{NQ}} & \multicolumn{3}{c}{\textbf{TriviaQA}} & \multicolumn{3}{c|}{\textbf{WebQ}} & \multicolumn{3}{c}{\textbf{NQ}} & \multicolumn{3}{c}{\textbf{TriviaQA}} & \multicolumn{3}{c}{\textbf{WebQ}} \\

                    &  & EM & Recall  & Con & EM & Recall  & Con & EM & Recall  & Con  & EM & Recall  & Con & EM & Recall  & Con & EM & Recall  & Con \\
\midrule
\multirow{4}{*}{BM25 } &       R  &    14.90  &  26.68  & 19.91  & 42.10 & 56.90 & 50.11 &  10.23 & 24.95 & 16.92 & 12.82 & 27.14 & 20.27 & 40.13 &58.40 &51.40 & 9.25 & 24.92 & 17.22 \\

&       G  &  24.68   &  45.43   &  33.85 &  52.23 &66.86 & 59.61 & 15.45  & 45.34 &35.28  &21.49  & 44.92 &33.65 & 48.74 &66.27 &  58.68 & 14.51& 45.73 & 35.03\\

&       R+G  & 25.51 & 45.22 & 34.29 & 53.29  & 70.04 & 62.75 & 15.05 & 42.75 & 32.77  & 22.29&45.46 & 34.37&  49.69 & 70.45&  62.98 & 13.97 & 43.91&33.56\\

&       G+R  &  25.67 & 45.93 &  34.70 & 53.24 & 69.54& 62.34 & 15.00 & 44.09 & 33.56  & 23.29 &46.10 &34.93 & 50.29 & 70.50 & 63.00& 13.43 & 43.53 & 32.72 \\

&   UPR   & 23.49  & 41.02 & 31.55 & 55.85 & 69.59 & 62.65 & 17.27 & 44.15 & 32.68 & 24.24 & 40.68 & 31.47 & 55.22 & 68.14 & 61.78 & 17.67 & 41.68 & 30.71 \\

&   RankGPT   & 28.45  & 42.73 & 34.02 & - & - & - & 19.73 & 40.57 & 29.82 & 26.65 & 40.89 & 32.44 & - & - & - & 18.55 & 37.92 & 28.10 \\

\midrule
\multirow{4}{*}{MSS} &       R  &  12.82 & 22.96 & 17.36 & 31.90 &44.27  & 37.75 &  7.38 & 18.9 &11.81  & 11.19 & 23.37 & 17.81 & 30.74 & 45.82 &38.91 & 6.88 & 19.30 & 12.40\\

&       G  & 24.95  &  45.82 & 34.15 &  51.69 &  66.58 & 59.22 & 15.84 &46.06 &35.77  &21.46 & 45.29 & 33.90 & 47.98& 65.57& 57.84 & 14.12 & 45.25 &45.25 \\

&       R+G  &  25.54 &   44.79 &33.96 & 51.39 &67.52 & 59.99 & 14.96 & 43.14 & 32.48 & 21.57 & 44.88 & 33.93 & 47.75& 68.22 & 60.41 & 14.46 & 44.46& 33.31 \\

&       G+R  &  25.31  & 45.34 & 34.43 & 51.81 &  67.72 & 60.12 &   15.20 &43.75  & 33.12 & 22.68 & 45.55 &34.68 & 48.75 & 68.02 & 60.27 & 13.43 & 44.14 &33.07\\

&   UPR   & 23.35  & 40.52 & 31.55 & 53.78 & 68.67 & 61.63 & 17.08 & 43.97 & 32.14 & 24.43 & 41.08 & 31.77 & 54.97 & 68.46 & 61.73 & 16.70 & 42.20 & 31.30 \\

&   RankGPT    & 28.17 & 41.67 & 33.21 & - & - & - & 19.05 & 39.33 & 29.58 & 25.84 & 39.07 & 31.41 & - & - & - & 17.37 & 36.38 & 27.21 \\

\midrule
\multirow{4}{*}{Contriever}   &       R  &  15.29  & 27.36 & 20.99 & 36.25 &  49.52 &  49.52 &  10.67 & 28.28 &  20.22 &13.24 & 27.68 &  21.88 &35.29  & 50.96 &44.19 & 9.35 &28.71 & 20.32\\

&       G  &  24.70 &  46.02 & 34.48 & 51.48  &  66.44 & 59.03 & 15.89  & 45.59 &35.23  &21.32 & 45.53 & 34.04 &  48.39 &65.82 &58.03 &  14.61 &45.67 &34.84\\

&       R+G  & 25.59  & 45.28 & 34.32 & 51.78 & 67.90 & 60.45 &   15.10 & 43.49 & 32.87 &22.18 & 45.26 & 34.45 & 48.38 & 68.84 & 61.47& 13.87 &45.07 &33.80\\

&       G+R  &  25.70 & 46.07   & 35.04  & 52.07 &  68.21& 60.98 & 15.50  &  44.08 & 33.21 & 23.13 &46.18  & 35.12 & 48.91 & 68.32 & 60.82 & 13.87 & 44.87 &34.10\\

&   UPR   & 23.24  & 41.01 & 31.61 & 53.48 & 68.70 & 61.82 & 17.32 & 44.11 & 32.97 & 24.21 & 40.99 & 31.63 & 55.71 & 68.74 & 61.98 & 17.62 & 37.73 & 28.44 \\

&   RankGPT     & 30.55  & 44.25 & 35.40 & - & - & - & 19.78 & 41.18 & 31.10 & 28.86 & 42.42 & 33.68 & - & - & - & 17.62 & 37.73 & 28.44 \\

\midrule
\multirow{4}{*}{DPR}  &       R  &   28.08 & 45.40 & 36.37 & 45.88 & 61.24 & 54.58 &  19.83 & 40.27   & 30.98  & 23.21 & 44.99 & 36.03 &43.62 & 62.61& 55.97 & 14.32 & 38.27 &28.98 \\

&       G  & 25.06 & 45.81 & 34.34 &  51.66  &  66.51 &59.08 &  15.45  & 46.32   & 36.59  &21.41 & 45.37 & 34.15 & 48.21&65.76 &57.96 & 14.76 & 45.32 &34.59\\

&       R+G  &  28.94 & 48.82 & 37.31 &   54.41 & 70.83 & 63.68 & 24.50 & 47.45  &  35.94 &24.62 & 48.50 & 37.28 & 50.61 & 71.30 & 63.99 & 14.32 & 46.55 & 34.94 \\

&       G+R  &  28.14 & 48.87 & 37.03 & 54.50& 70.53 & 63.42 &25.51 & 48.60  &  37.50&25.51 & 50.28 & 38.53 & 51.65 & 71.66 &64.30 &14.81 & 46.59& 35.23\\

&   UPR   & 23.60  & 41.18 & 31.77 & 53.41 & 68.60 & 61.60 & 18.06 & 44.01 & 32.48 & 24.74 & 41.32 & 31.75 & 56.07 & 69.14 & 62.41 & 19.64 & 39.77 & 29.87 \\

&   RankGPT    & 31.74 & 46.41 & 36.76 & - & - & - & 20.42 & 40.84 & 31.00 & 29.58 & 43.92 & 34.65 & - & - & - & 19.64 & 39.77 & 29.87 \\

\midrule
\multirow{4}{*}{MSS+DPR}&       R  &   28.17 & 46.72 & 37.00 &  47.69 & 63.66 & 57.08 &  13.92 &  40.87   & 30.57 & 23.68 & 46.70 &37.53 & 45.72 & 65.24 & 58.53 & 13.92 &38.97 & 29.67\\

&       G  &  24.73  &  45.46 & 33.85 & 51.64  &66.90 & 59.40&  14.51  & 48.74   &  37.73 & 21.80 & 45.28 & 33.90 &  47.98 & 65.51 & 57.83 &14.27 &46.14 &35.03\\

&       R+G  &  29.41 &   49.73 & 38.25 &   54.53 &70.86  &63.82 & 14.96 & 49.48  & 38.84 &24.09 & 48.80 & 37.34 & 50.72 & 71.67 & 64.35 & 14.96 &46.40 & 34.94\\

&       G+R  &  28.61 & 49.27 & 37.36 &  54.48 & 71.16  & 63.91& 14.76 &  48.80  & 37.77 & 25.54 &50.30 &38.61 & 52.20 & 72.19 & 64.86 & 14.76 &46.42 & 35.33\\

&   UPR   & 23.10  & 40.81 & 31.25 & 53.65 & 68.61 & 61.69 & 16.68 & 44.07 & 32.48 & 25.24 & 41.17 & 31.83 & 55.52 & 68.76 & 61.90 & 18.99 & 40.76 & 30.22 \\

&   RankGPT   & 31.94 & 45.95 & 36.59 & - & - & - & 21.41 & 42.54 & 32.14 & 29.95 & 45.00 & 36.49 & - & - & - & 18.99 & 40.76 & 30.22 \\

\bottomrule
\end{tabular}
}
\caption{Zero-shot results of in-context learning on
The test set of NQ, TriviaQA, and WebQ uses the LLama 3/3.1 8B Model as RAG }
\label{tab:qa_LLama_3_3.1}
\end{table*}

\begin{table*}[!ht]
\centering
\resizebox{1.0\textwidth}{!}{  
\setlength\tabcolsep{3pt}
\begin{tabular}{@{}l|c | ccc| ccc | ccc |ccc | ccc | ccc @{}}
\toprule
 & & \multicolumn{9}{c|}{\textbf{Gemma-2-2b}} & \multicolumn{9}{c}{\textbf{Gemma-2-9b}}\\

\multirow{3}{*}{\textbf{Retriever} }& \multirow{3}{*}{\textbf{\# Doc}} & \multicolumn{3}{c}{\textbf{NQ}} & \multicolumn{3}{c}{\textbf{TriviaQA}} & \multicolumn{3}{c|}{\textbf{WebQ}} & \multicolumn{3}{c}{\textbf{NQ}} & \multicolumn{3}{c}{\textbf{TriviaQA}} & \multicolumn{3}{c}{\textbf{WebQ}} \\

&  & EM & Recall  & Con & EM & Recall  & Con & EM & Recall  & Con  & EM & Recall  & Con & EM & Recall  & Con & EM & Recall  & Con \\
\midrule

\multirow{6}{*}{BM25 } &       R   & 14.02 & 25.55 & 18.53 & 43.28 & 52.78 & 47.01 & 14.71 & 37.74 & 27.21 & 19.81 & 26.95 & 22.05 & 57.55 & 65.53 & 60.29 & 14.96 & 24.87 & 20.13 \\

&       G   & 27.01 & 46.85 & 35.35 & 59.91 & 72.99 & 66.36 & 19.34 & 50.19 & 38.93 & 28.28 & 46.46 & 36.04 & 63.02 & 75.33 & 68.83 & 18.65 & 50.57 & 39.37 \\

&       R+G   & 28.39 & 46.41 & 35.62 & 59.89 & 73.39 & 66.53 & 19.29 & 47.44 & 35.93 & 28.45 & 45.09 & 35.54 & 63.50 & 75.23 & 68.50 & 19.05 & 45.38 & 34.94 \\

&       G+R   & 28.50 & 46.50 & 35.68 & 59.87 & 72.99 & 66.21 & 19.73 & 48.22 & 36.47 & 28.42 & 45.67 & 35.96 & 62.94 & 75.44 & 68.64 & 19.34 & 45.83 & 35.53 \\

&   UPR   & 26.23 & 44.37 & 33.99 & 58.71 & 71.59 & 64.79 & 19.78 & 48.16 & 36.47 & 23.41 & 41.52 & 32.24 & 58.74 & 71.92 & 65.20 & 15.94 & 46.16 & 34.15 \\

&   RankGPT   & 30.36 & 45.96 & 36.09 & - & - & - & 21.11 & 45.13 & 34.35 & 30.75 & 43.48 & 35.46 & - & - & - & 21.06 & 39.32 & 31.15 \\

\midrule
\multirow{6}{*}{MSS} &       R   & 13.96 & 25.41 & 18.50 & 33.05 & 42.24 & 36.23 & 14.71 & 37.74 & 27.21 & 19.78 & 26.86 & 22.08 & 50.93 & 58.57 & 53.28 & 14.96 & 24.92 & 20.18 \\

&       G   & 27.06 & 46.87 & 35.29 & 59.27 & 72.32 & 65.49 & 19.34 & 50.19 & 38.93 & 27.95 & 46.77 & 36.01 & 62.67 & 74.93 & 68.41 & 18.65 & 50.57 & 39.37 \\

&       R+G   & 28.48 & 46.36 & 35.48 & 58.63 & 71.48 & 64.65 & 19.29 & 47.05 & 35.78 & 28.56 & 45.27 & 35.65 & 62.53 & 73.97 & 67.27 & 18.80 & 44.44 & 34.01 \\

&       G+R   & 28.42 & 46.18 & 35.57 & 58.47 & 71.27 & 64.46 & 19.78 & 47.79 & 36.07 & 28.31 & 45.53 & 35.37 & 61.78 & 73.76 & 67.07 & 18.55 & 45.11 & 35.09 \\

&   UPR   & 26.20 & 44.14 & 33.77 & 58.15 & 71.02 & 64.10 & 19.73 & 48.92 & 36.96 & 23.10 & 41.49 & 32.05 & 57.50 & 71.15 & 64.33 & 15.85 & 47.20 & 34.65 \\

&   RankGPT   & 29.17 & 45.00 & 35.32 & - & - & - & 19.93 & 44.61 & 34.10 & 29.97 & 42.59 & 34.85 & - & - & - & 19.64 & 36.91 & 29.72 \\

\midrule
\multirow{6}{*}{Contriever}   &       R   & 13.96 & 25.41 & 18.50 & 33.05 & 42.24 & 36.23 & 14.71 & 37.74 & 27.21 & 19.78 & 26.86 & 22.08 & 50.93 & 58.57 & 53.28 & 14.96 & 24.92 & 20.18 \\

&       G   & 27.06 & 46.87 & 35.29 & 59.27 & 72.32 & 65.49 & 19.34 & 50.19 & 38.93 & 27.95 & 46.77 & 36.01 & 62.67 & 74.93 & 68.41 & 18.65 & 50.57 & 39.37 \\

&       R+G   & 28.78 & 46.98 & 36.12 & 58.86 & 71.68 & 64.81 & 20.28 & 48.42 & 37.30 & 28.84 & 45.33 & 35.90 & 62.85 & 73.89 & 67.21 & 19.64 & 46.01 & 35.43 \\

&       G+R   & 28.75 & 46.70 & 35.98 & 58.64 & 71.33 & 64.57 & 20.13 & 48.24 & 36.86 & 28.37 & 45.46 & 36.07 & 61.97 & 73.87 & 67.07 & 19.09 & 45.95 & 36.17 \\

&   UPR   & 26.26 & 44.26 & 33.82 & 58.37 & 71.17 & 64.23 & 19.83 & 48.32 & 36.61 & 23.21 & 41.46 & 32.08 & 57.64 & 71.20 & 64.36 & 16.14 & 46.12 & 34.01 \\

&   RankGPT   & 32.11 & 47.86 & 38.31 & - & - & - & 20.67 & 45.77 & 34.89 & 32.44 & 44.83 & 36.79 & - & - & - & 19.88 & 38.57 & 30.02 \\

\midrule
\multirow{6}{*}{DPR}  &       R   & 13.99 & 25.44 & 18.53 & 33.05 & 42.24 & 36.23 & 14.71 & 37.74 & 27.21 & 19.78 & 26.86 & 22.08 & 50.93 & 58.57 & 53.28 & 14.96 & 24.92 & 20.18 \\

&       G  & 27.06 & 46.87 & 35.29 & 59.27 & 72.32 & 65.49 & 19.34 & 50.19 & 38.93 & 27.92 & 46.75 & 35.98 & 62.67 & 74.93 & 68.41 & 18.65 & 50.57 & 39.37 \\

&       R+G   & 30.25 & 48.89 & 37.73 & 60.16 & 73.19 & 66.28 & 20.18 & 48.36 & 36.66 & 30.83 & 47.93 & 37.89 & 63.72 & 74.99 & 68.27 & 19.93 & 46.16 & 35.88 \\

&       G+R   & 30.72 & 48.78 & 37.45 & 60.17 & 72.98 & 66.22 & 20.52 & 48.93 & 36.86 & 29.92 & 47.48 & 37.12 & 63.18 & 75.00 & 68.29 & 19.39 & 46.59 & 36.32 \\

&   UPR   & 26.51 & 44.62 & 34.16 & 58.71 & 71.72 & 64.78 & 19.78 & 48.62 & 36.86 & 23.38 & 41.80 & 32.38 & 57.85 & 71.54 & 64.75 & 16.04 & 46.56 & 34.50 \\

&   RankGPT   & 34.16 & 50.26 & 39.42 & - & - & - & 21.21 & 46.41 & 35.48 & 34.04 & 47.14 & 38.64 & - & - & - & 21.01 & 40.41 & 31.69 \\

\midrule
\multirow{6}{*}{MSS+DPR}&       R   & 13.96 & 25.41 & 18.50 & 33.05 & 42.24 & 36.23 & 14.71 & 37.74 & 27.21 & 19.78 & 26.86 & 22.08 & 50.93 & 58.57 & 53.28 & 14.96 & 24.92 & 20.18 \\

&       G   & 27.06 & 46.87 & 35.29 & 59.27 & 72.32 & 65.49 & 19.34 & 50.19 & 38.93 & 28.25 & 46.47 & 36.01 & 62.67 & 74.93 & 68.41 & 18.60 & 46.85 & 36.32 \\

&       R+G   & 30.91 & 49.23 & 38.12 & 60.56 & 73.67 & 66.82 & 20.72 & 48.67 & 36.91 & 30.58 & 47.85 & 37.73 & 63.76 & 75.20 & 68.48 & 19.93 & 46.62 & 36.32 \\

&       G+R  & 30.66 & 49.15 & 37.98 & 60.25 & 73.53 & 66.68 & 21.26 & 49.67 & 37.70 & 29.94 & 47.61 & 37.40 & 63.17 & 75.19 & 68.37 & 20.13 & 46.90 & 37.01 \\

&   UPR   & 26.54 & 44.67 & 34.21 & 58.64 & 71.50 & 64.55 & 19.49 & 48.03 & 36.02 & 23.46 & 41.75 & 32.38 & 57.75 & 71.34 & 64.48 & 15.94 & 45.96 & 33.56 \\

&   RankGPT   & 32.51 & 49.77 & 39.53 & - & - & - & 21.41 & 47.58 & 36.52 & 32.13 & 46.55 & 38.29 & - & - & - & 21.06 & 40.72 & 32.04 \\

\bottomrule
\end{tabular}
}
\caption{
Zero-shot results of in-context learning on
The test set of NQ, TriviaQA, and WebQ uses the Gemma Model as RAG.
}
\label{tab:qa_Gemma}
\end{table*}
\begin{table*}[!ht]
\centering
\resizebox{1.0\textwidth}{!}{  
\setlength\tabcolsep{3pt}
\begin{tabular}{@{}l|c | ccc| ccc | ccc |ccc | ccc | ccc @{}}
\toprule
 & & \multicolumn{9}{c|}{\textbf{Llama-2-13b-hf}} & \multicolumn{9}{c}{\textbf{Mistral-7B-v0.1}}\\

\multirow{3}{*}{\textbf{Retriever} }& \multirow{3}{*}{\textbf{\# Mode}} & \multicolumn{3}{c}{\textbf{NQ}} & \multicolumn{3}{c}{\textbf{TriviaQA}} & \multicolumn{3}{c|}{\textbf{WebQ}} & \multicolumn{3}{c}{\textbf{NQ}} & \multicolumn{3}{c}{\textbf{TriviaQA}} & \multicolumn{3}{c}{\textbf{WebQ}} \\

&  & EM & Recall  & Con & EM & Recall  & Con & EM & Recall  & Con  & EM & Recall  & Con & EM & Recall  & Con & EM & Recall  & Con \\
\midrule

\multirow{6}{*}{BM25 } 
&       R   & 21.14 & 30.82 & 24.46 & 57.90 & 65.27 & 59.57 & 19.54 & 37.38 & 27.51 & 11.19 & 13.45 & 11.80 & 52.85 & 58.11 & 53.82 & 6.40 & 8.46 & 6.84 \\

&       G   & 28.06 & 44.60 & 34.21 & 62.64 & 73.63 & 67.13 & 20.32 & 45.54 & 34.40 & 27.01 & 41.30 & 32.19 & 62.64 & 72.63 & 66.19 & 16.09 & 33.01 & 24.70 \\

&       R+G   & 26.62 & 41.96 & 32.47 & 61.35 & 73.01 & 66.34 & 19.00 & 43.61 & 32.53 & 25.68 & 37.68 & 30.11 & 60.45 & 69.76 & 63.49 & 15.65 & 29.42 & 22.54 \\

&       G+R   & 26.79 & 43.16 & 33.35 & 62.01 & 73.14 & 66.67 & 19.24 & 43.70 & 32.68 & 23.71 & 34.64 & 27.87 & 58.56 & 67.61 & 61.63 & 13.44 & 26.74 & 20.08 \\

&   UPR   & 27.59 & 42.65 & 32.99 & 61.60 & 71.62 & 65.00 & 20.37 & 44.09 & 33.76 & 25.18 & 40.47 & 31.11 & 59.64 & 69.91 & 63.20 & 17.18 & 40.89 & 30.46 \\

&   RankGPT   & 29.22 & 43.00 & 34.21 & - & - & - & 21.99 & 41.25 & 31.15 & 25.35 & 40.47 & 31.36 & - & - & - & 17.18 & 40.90 & 30.46 \\

\midrule
\multirow{6}{*}{MSS} 
&       R   & 21.52 & 30.92 & 24.18 & 51.75 & 58.58 & 53.35 & 20.62 & 39.45 & 29.68 & 11.08 & 13.33 & 11.66 & 42.69 & 47.12 & 43.40 & 6.40 & 8.46 & 6.84 \\

&       G   & 28.01 & 43.22 & 33.74 & 60.66 & 72.01 & 65.43 & 19.44 & 44.80 & 34.20 & 27.15 & 41.40 & 32.19 & 61.28 & 71.06 & 64.71 & 16.09 & 33.01 & 24.70 \\

&       R+G   & 26.81 & 42.48 & 33.02 & 58.64 & 70.39 & 63.87 & 19.05 & 42.74 & 32.14 & 25.90 & 38.34 & 30.58 & 56.28 & 64.82 & 59.03 & 14.57 & 27.91 & 20.96 \\

&       G+R   & 27.48 & 43.97 & 33.88 & 59.56 & 71.14 & 64.63 & 18.75 & 42.48 & 31.64 & 23.10 & 33.75 & 27.26 & 52.15 & 59.83 & 54.59 & 13.09 & 24.77 & 18.31 \\

&   UPR   & 27.29 & 42.28 & 32.63 & 59.23 & 69.93 & 63.56 & 20.72 & 44.50 & 33.96 & 23.77 & 40.41 & 30.97 & 57.24 & 67.67 & 61.04 & 17.42 & 42.17 & 31.20 \\

&   RankGPT   & 27.56 & 41.77 & 33.05 & - & - & - & 20.77 & 44.50 & 33.96 & 23.77 & 39.08 & 29.28 & - & - & - & 16.98 & 39.87 & 29.43 \\

\midrule
\multirow{6}{*}{Contriever}   
&       R   & 20.47 & 30.13 & 23.85 & 42.69 & 47.12 & 43.40 & 19.98 & 37.74 & 27.61 & 11.08 & 13.33 & 11.66 & 42.69 & 47.12 & 43.40 & 6.40 & 8.46 & 6.84 \\

&       G   & 26.79 & 43.16 & 33.35 & 61.28 & 71.06 & 64.71 & 19.69 & 44.88 & 33.76 & 27.15 & 41.40 & 32.19 & 61.28 & 71.06 & 64.71 & 16.09 & 33.01 & 24.70 \\

&       R+G   & 27.45 & 42.89 & 33.27 & 56.43 & 65.00 & 59.20 & 19.59 & 43.25 & 32.19 & 25.35 & 37.57 & 30.08 & 56.43 & 65.00 & 59.20 & 15.31 & 27.88 & 20.82 \\

&       G+R   & 27.17 & 43.64 & 33.74 & 52.66 & 60.52 & 55.33 & 19.88 & 44.21 & 33.61 & 23.02 & 33.53 & 27.29 & 54.39 & 62.52 & 57.18 & 13.44 & 25.79 & 19.24 \\

&   UPR   & 26.57 & 41.99 & 32.44 & 59.60 & 70.02 & 63.58 & 20.72 & 44.65 & 33.56 & 25.10 & 40.43 & 31.02 & 57.26 & 67.63 & 61.11 & 17.27 & 40.48 & 29.97 \\

&   RankGPT   & 30.39 & 44.60 & 36.09 & - & - & - & 20.72 & 44.65 & 33.56 & 25.10 & 40.48 & 31.02 & - & - & - & 17.27 & 40.48 & 29.97 \\

\midrule
\multirow{6}{*}{DPR}  
&       R   & 21.94 & 31.36 & 24.88 & 51.07 & 57.97 & 52.71 & 19.83 & 37.35 & 28.05 & 11.11 & 13.36 & 11.69 & 42.69 & 47.12 & 43.40 & 6.40 & 8.46 & 6.84 \\

&       G  & 28.12 & 44.10 & 34.32 & 60.85 & 71.87 & 65.44 & 20.47 & 45.05 & 34.15 & 27.15 & 41.40 & 32.19 & 61.28 & 71.06 & 64.71 & 16.09 & 33.01 & 24.70 \\

&       R+G   & 28.81 & 44.73 & 34.82 & 58.00 & 67.03 & 61.14 & 20.32 & 45.14 & 34.25 & 27.70 & 40.40 & 32.47 & 58.00 & 67.03 & 61.14 & 16.44 & 32.14 & 24.31 \\

&       G+R   & 27.92 & 44.15 & 34.27 & 60.01 & 71.44 & 64.93 & 20.72 & 45.02 & 34.15 & 25.01 & 35.90 & 29.06 & 54.64 & 62.57 & 57.27 & 14.71 & 28.39 & 21.75 \\

&   UPR   & 27.45 & 42.93 & 33.02 & 60.05 & 70.49 & 63.96 & 20.62 & 44.81 & 33.91 & 25.26 & 40.71 & 31.36 & 57.53 & 67.80 & 61.31 & 17.18 & 41.20 & 30.66 \\

&   RankGPT   & 32.77 & 47.56 & 38.06 & - & - & - & 22.15 & 44.37 & 33.81 & 25.26 & 40.71 & 31.36 & - & - & - & 17.18 & 41.20 & 30.66 \\

\midrule
\multirow{6}{*}{MSS+DPR}
&       R   & 21.47 & 31.26 & 24.60 & 51.35 & 58.26 & 53.01 & 19.83 & 37.37 & 27.61 & 11.08 & 13.33 & 11.66 & 42.69 & 47.12 & 43.40 & 6.40 & 8.46 & 6.84 \\

&       G   & 28.20 & 43.75 & 34.07 & 60.44 & 71.60 & 65.09 & 20.13 & 45.50 & 34.30 & 27.15 & 41.40 & 32.19 & 61.28 & 71.06 & 64.71 & 16.09 & 33.01 & 24.70 \\

&       R+G   & 28.45 & 45.21 & 35.21 & 59.50 & 71.17 & 64.79 & 19.78 & 44.57 & 33.46 & 27.40 & 39.95 & 31.94 & 58.03 & 67.43 & 61.49 & 16.44 & 31.62 & 23.97 \\

&       G+R  & 28.73 & 44.67 & 34.79 & 59.96 & 71.90 & 65.57 & 20.77 & 45.09 & 34.10 & 25.07 & 35.95 & 28.92 & 54.39 & 62.52 & 57.18 & 14.96 & 29.13 & 22.00 \\

&   UPR   & 27.45 & 42.91 & 33.10 & 60.00 & 71.52 & 64.07 & 19.98 & 44.81 & 33.32 & 25.24 & 40.59 & 31.25 & 57.47 & 67.80 & 61.12 & 16.09 & 44.33 & 33.32 \\

&   RankGPT   & 32.61 & 48.90 & 40.19 & - & - & - & 21.95 & 43.73 & 33.32 & 25.24 & 40.59 & 31.25 & - & - & - & 16.09 & 44.33 & 33.32 \\

\bottomrule
\end{tabular}
}
\caption{
Zero-shot results of in-context learning on
the test set of NQ, TriviaQA, and WebQ  using Llama-2-13b-hf and Mistral-7B-v0.1 as RAG
}
\label{tab:qa_Llama-2-13b}
\end{table*}

\end{document}